\def\paperID{3439}
\def\confName{CVPR}
\def\confYear{2024}

\def\paperTitle{\textit{PEGASUS}: Personalized Generative 3D Avatars with Composable Attributes}

\def\authorBlock{
    Hyunsoo Cha \qquad
    Byungjun Kim \qquad
    Hanbyul Joo \\
    Seoul National University\\
    {\tt\small 243stephen@snu.ac.kr \quad byungjun.kim@snu.ac.kr \quad hbjoo@snu.ac.kr}\\
    {\tt\small \href{https://snuvclab.github.io/pegasus/}{\color{magenta}{https://snuvclab.github.io/pegasus/}}}
    \vspace{-5mm}
}

\newif\ifreview 
\newif\ifarxiv \newcommand{\arxiv}{\arxivtrue}
\newif\ifcamera 
\newif\ifrebuttal

\documentclass[10pt,twocolumn,letterpaper]{article}
\arxiv
\usepackage[pagenumbers]{cvpr} %

\usepackage[dvipsnames]{xcolor}

\usepackage{cuted} %

\title{\paperTitle}
\author{\authorBlock}

\ifreview \usepackage[review]{cvpr} \fi
\ifarxiv \usepackage[pagenumbers]{cvpr} \fi
\ifrebuttal \usepackage[rebuttal]{cvpr} \fi
\ifcamera \usepackage{cvpr} \fi

\usepackage{graphicx}
\usepackage{amsmath}
\usepackage{amssymb}
\usepackage{booktabs}

\usepackage{times}
\usepackage{microtype}
\usepackage{epsfig}
\usepackage{xcolor}
\usepackage{caption}
\usepackage{float}
\usepackage{placeins}
\usepackage{color, colortbl}
\usepackage{stfloats}
\usepackage{enumitem}
\usepackage{tabularx}
\usepackage{xstring}
\usepackage{multirow}
\usepackage{xspace}
\usepackage{cuted} %
\usepackage{url}
\usepackage{subcaption}
\usepackage{xcolor}
\usepackage[hang,flushmargin]{footmisc}
\usepackage{bm}
\usepackage{amsmath}
\usepackage{booktabs} %
\usepackage{makecell} %
\usepackage{amssymb}
\usepackage{indentfirst}
\ifcamera \usepackage[accsupp]{axessibility} \fi
\ifcamera  \fi
\newcommand{\ourmethod}[1]{OFHR}

\ifarxiv  \fi

\renewcommand{\vec}[1]{\bm{#1}}

\newcommand{\R}[1]{{%
    \textbf{%
        \ifstrequal{#1}{1}{\textcolor{orange}{R#1}}{%
        \ifstrequal{#1}{2}{\textcolor{blue}{R#1}}{%
        \ifstrequal{#1}{3}{\textcolor{magenta}{R#1}}{%
        \ifstrequal{#1}{4}{\textcolor{teal}{R#1}}{%
                           \textcolor{cyan}{R#1}%
        }}}}%
    }%
}}

\newcommand{\figref}[1]{Fig.~\ref{#1}}
\newcommand{\tabref}[1]{Tab.~\ref{#1}}
\newcommand{\eqnref}[1]{Eq.~(\ref{#1})}
\newcommand{\secref}[1]{Sec.~\ref{#1}}

\usepackage{subcaption} %

\usepackage{xr-hyper}

\makeatletter
\newcommand*{\addFileDependency}[1]{
  \typeout{(#1)}
  \@addtofilelist{#1}
  \IfFileExists{#1}{}{\typeout{No file #1.}}
}

\makeatother

\usepackage[pagebackref,breaklinks,colorlinks]{hyperref}
\usepackage[capitalize]{cleveref}
\crefname{section}{Sec.}{Secs.}
\crefname{table}{Table}{Tables}
\crefname{figure}{Fig.}{Figs.}

\begin{document}
\maketitle
\begin{strip}\centering
\includegraphics[width=\linewidth, trim={0 0cm 0 0.0cm}, clip]{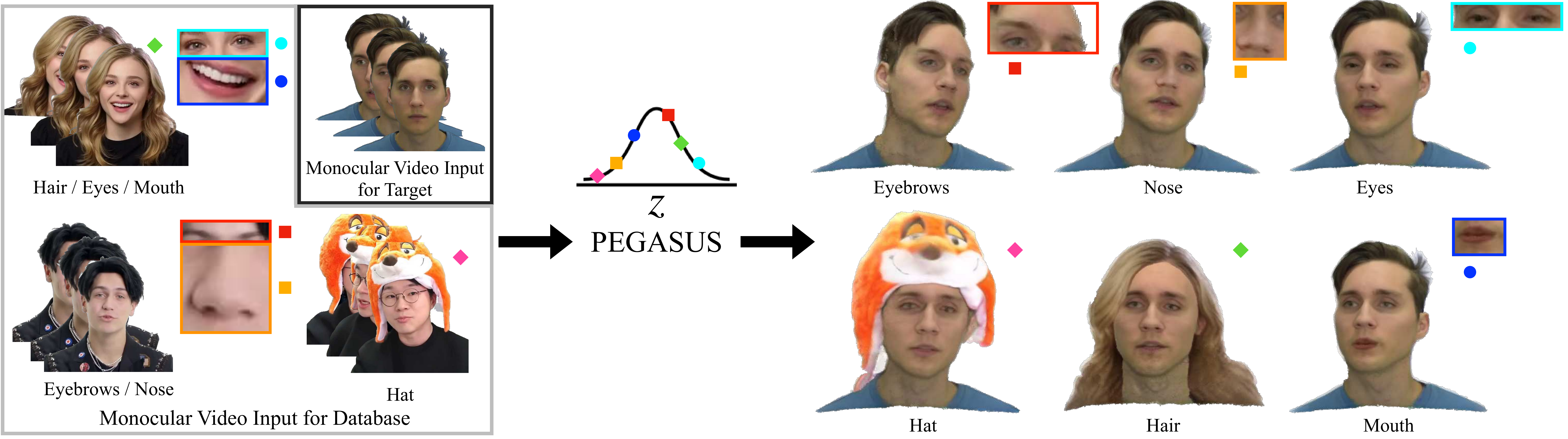} %
\captionof{figure}{\textbf{PEGASUS.} Our method builds a personalized generative 3D face avatar from monocular video sources.}
\label{fig:teaser}
\end{strip}
\begin{abstract}
We present \emph{PEGASUS}, a method for constructing a personalized generative 3D face avatar from monocular video sources. Our generative 3D avatar enables disentangled controls to selectively alter the facial attributes (e.g., hair or nose) while preserving the identity. Our approach consists of two stages: synthetic database generation and constructing a personalized generative avatar. We generate a synthetic video collection of the target identity with varying facial attributes, where the videos are synthesized by borrowing the attributes from monocular videos of diverse identities. Then, we build a person-specific generative 3D avatar that can modify its attributes continuously while preserving its identity. Through extensive experiments, we demonstrate that our method of generating a synthetic database and creating a 3D generative avatar is the most effective in preserving identity while achieving high realism. Subsequently, we introduce a zero-shot approach to achieve the same goal of generative modeling more efficiently by leveraging a previously constructed personalized generative model.
\end{abstract}
\vspace{-10px}

\section{Introduction}
\label{sec:intro}
Building a personalized 3D avatar for representing an individual in virtual spaces can bring significant advancements in the field of AR/VR and applications within the metaverse. Importantly, the method should be user-friendly to allow novices to build their avatars without the need for complex capture systems. It should also offer a high level of realism, depicting the fine-grained details of the individual's geometry and appearance, and, importantly, the avatar should be animatable to mirror the user's facial expressions in the virtual space. However, the 3D avatar does not need to maintain an exact replica of the user's single appearance, as users may prefer to alter their avatars. This includes modifications of changing hairstyles, adding accessories like hats, or even altering facial parts to give the avatar a more aesthetically pleasing look, such as adopting the appearance of celebrities.

Recent technologies make it possible to build high-quality 3D face avatars for general users from monocular video inputs only~\cite{guo2023vid2avatar, zheng2023pointavatar, zheng2022imavatar, bharadwaj2023flare, grassal2022neural, gafni2021dynamic}. By leveraging parametric morphable face models~\cite{li2017flame, blanz19993dmm}, these approaches produce realistic animatable human avatars from sparse monocular videos that capture naturally moving facial images, by fusing observed cues into a canonical space. However, the previous approaches mainly focus on creating the exact replica from the input videos, without providing the functionality to alter the subparts of the avatars, such as hairstyles or nose. As an alternative direction, generative models in producing realistic faces have been studied in the 2D field, by producing 2D human faces with diverse appearance changes and facial expression changes~\cite{xu2022transeditor, shi2022semanticstylegan, zhu2021barbershop}. 3D-aware generative models leveraging the pre-trained 2D generative models are also presented for generative face modeling in 3D~\cite{chan2022efficient, chang2023hairnerf}. While this approach shows realistic faces, they are not fully animatable, lacking explicit mapping to the 3D morphable models, and, thus, it is difficult to reenact the facial expressions from the target or allow viewpoint changes while keeping the identity.

In this work, we present PEGASUS, a method to build a \emph{personalized generative 3D avatar} from monocular video inputs. In contrast to the previous work~\cite{zheng2022imavatar, zheng2023pointavatar}, our 3D avatar enables compositional controls of facial attributes, where users can make alterations for desired facial attributes such as hair, nose, or accessories, as shown in ~\figref{fig:teaser}, while preserving the identity of the target person. The control can be performed by changing the disentangled latent codes defined in a continuous latent space.
Our personalized generative 3D avatar is constructed from the monocular video of the target individual. Importantly, to learn the possible variations of each facial attribute, we leverage other available monocular videos from arbitrary individuals, where our personalized generative models can automatically learn continuous disentangled latent spaces of facial attributes. 

However, there exist significant challenges in consolidating the monocular videos from multiple individuals into a personalized generative model construction for the target individual. Building a model with videos from many individuals often fails to preserve the fine-grained appearance details of the target individual, and, more critically, changing the latent space can lead to changes in the entire facial appearance, rather than selectively altering the desired subpart. 
As a solution, we present an approach by synthesizing part-swapped videos of the target individual by replacing a specific facial part with the one from other individuals, as shown in \figref{fig:synthesized_videos}.
Built with diverse part-swapped videos, our generative 3D avatar, PEGASUS, can preserve high-quality details for the target individuals, while equipped with the generative power to selectively alter each facial part. While our generative 3D avatar already shows satisfactory performance, it involves the time-consuming process of constructing a set of part-swapped videos. As a more rapid and efficient solution, we further introduce an approach that achieves the same objectives through \emph{zero-shot part transfer}, leveraging previously constructed personalized generative models. Through several experiments, we demonstrate the superior performance of our approach when compared to alternative methods.

Our contributions are summarized as follows:
(1) the first method to build personalized generative 3D avatars from monocular video sources; (2) disentangled controllability to selectively alter a subpart or multiple parts of the 3D faces of the target individual; and (3) the 3D part transfer approach to efficiently implement personalized generative models without additional training.

\section{Related Work}
\label{sec:related}
\vspace{-5px}
\noindent \textbf{3D Facial Avatar Reconstruction.}
To deal with the inherent diversity and dynamics of human faces, 3D parametric face models have been proposed to represent 3D facial changes via a set of parameters that model variations of the shapes, poses, and expressions of faces~\cite{blanz19993dmm, cao2013facewarehouse, li2017flame}. DECA~\cite{feng2021learning} proposes a method for regressing FLAME~\cite{li2017flame} parameters from monocular images, which enables 3D facial avatar reconstruction without a 3D scan setup. With the advancements of neural rendering~\cite{mildenhall2021nerf}, several neural rendering-based 3D facial avatar reconstruction approaches were proposed to overcome the limited facial details of the parametric model~\cite{feng2021learning, danvevcek2022emoca, zielonka2022towards, sanyal2019learning, gafni2021dynamic, grassal2022neural, zheng2022imavatar, zheng2023pointavatar, xu2023latentavatar, bai2023high}. IMAvatar~\cite{zheng2022imavatar} introduces the dynamic 3D morphable face model as an implicit representation with canonicalization of the head deformation and facial expression based on SNARF~\cite{chen2021snarf}. 
PointAvatar~\cite{zheng2023pointavatar} proposes a deformable point-based representation to reconstruct high-frequency details from monocular video with efficient rendering. 

\noindent \textbf{Face Editing in 2D/3D.}
GAN~\cite{goodfellow2014generative} has made significant contributions to producing natural and high-quality face editing ~\cite{xu2022transeditor, shi2022semanticstylegan, yin2022styleheat, zhang2022training, zhu2021barbershop, kim2022style, chang2023hairnerf, zhu2022hairnet, zhu2020domain, chan2022efficient}. 
For instance, Barbershop~\cite{zhu2021barbershop} edits the hairstyle of the target face to match the source appearance by manipulating the latent space that represents the feature's spatial location and appearance.
With the rise of the diffusion model~\cite{ho2020denoising, song2020denoising}, methods for editing faces or scenes based on diffusion have been proposed~\cite{meng2021sdedit, hertz2022prompt, brooks2023instructpix2pix}. For instance, Instruct-Pix2Pix~\cite{brooks2023instructpix2pix} introduces a method for editing an input image with text instructions by finetuning a pretrained diffusion model~\cite{rombach2022high} using a generated image editing dataset~\cite{hertz2022prompt} through supervised learning. Following the emergence of 2D diffusion models, several approaches leverage the pretrained diffusion models to edit 3D scenes from the text prompts~\cite{haque2023instruct, zhuang2023dreameditor, sella2023vox}.
For instance, Instruct-NeRF2NeRF~\cite{haque2023instruct} proposes an iterative optimization process for editing a pretrained NeRF scene~\cite{mildenhall2021nerf}.
However, these approaches struggle to preserve the identity of the scene and require additional optimization.

\noindent \textbf{Compositional Modeling for 3D Face Avatar.}
Several methods propose to edit implicit representations leveraging learnable or pretrained latent spaces~\cite{yenamandra2021i3dmm, kim20233d, jiang2022nerffaceediting, ho2023learning}. 
However, due to the difficulty in editing, recent approaches introduce decoupled representations for garments or attributes~\cite{li2023megane, kim2023ncho, feng2022capturing, feng2023learning, ho2023learning}. 
MEGANE~\cite{li2023megane} attaches reconstructed eyeglasses from 3D scans to volumetric primitive 3D avatars, optimizing deformation for avatar-specific adjustments. %
SCARF and DELTA~\cite{feng2022capturing, feng2023learning} create 3D avatars using a hybrid representation, enabling the transfer of garments or hair without additional optimization. 
However, these approaches are limited to specific categories or attributes, such as eyeglasses and bags, and the synthesized result avatars are not natural. Furthermore, they lack the ability to change or generate facial attributes continuously.

\begin{figure*}[ht]\centering
\includegraphics[width=\linewidth, trim={0 0 0 0},clip]{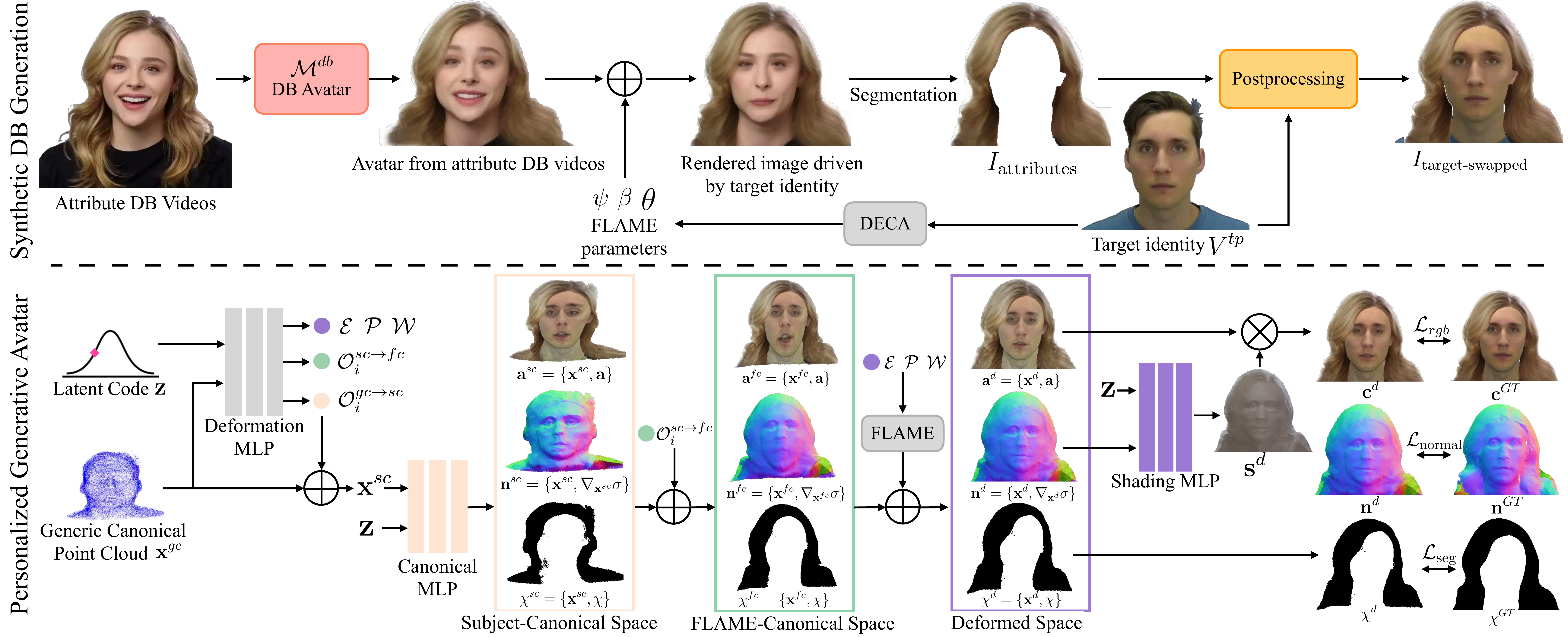}
\vspace{-15px}
\captionof{figure}{\textbf{Method Overview.} 
Our approach consists of two main components: synthetic database (DB) generation and a personalized generative 3D avatar model. Initially, we build a synthetic DB via face part swapping from the attribute DB videos. For the generation of the synthetic DB, we propose the method through post-processing and attribute alignment leveraging FLAME parameters. Subsequently, we train our model utilizing the synthetic DB that contains the same target identity with varying attributes. Our model infers the 3D point locations in the deformed space $\mathbf{x}^d$, normal $\mathbf{n}^d$, shading $\mathbf{s}^d$, point segment cues $\chi^d$, and the albedo $\mathbf{a}^d$ for each queried canonical point $\mathbf{x}^{gc}$, conditioned by the latent code $\mathbf{z}$.}
\label{fig:overview}
\vspace{-5px}
\end{figure*}

\section{Preliminaries: PointAvatar~\cite{zheng2023pointavatar}} \label{sec:preliminaries}
\vspace{-3px}
Our approach extends PointAvatar, which reconstructs the 3D avatar of a single identity from a monocular video, to a personalized generative avatar with controllable facial attributes. PointAvatar represents the target avatar via the initial canonical learnable point representations $P_c = \{x^c_i\}_{i = \{ 1\cdots N\}}$, where $x^c_i \in \mathbb{R}^3$ represents $i$-th learnable point defined in the canonical space (denoted as the superscript $c$). By estimating the offset value $ \mathcal{O}_i^{c\rightarrow fc}$ from a trained MLP, the canonical points are deformed into the \emph{FLAME-canonical space} (denoted as $fc$) as: $\mathbf{x}^{fc}_i = \mathbf{x}^{c}_i + \mathcal{O}_i^{c\rightarrow fc}$. Subsequently,  the points are deformed into the posed space as leveraging FLAME model~\cite{li2017flame}:
\vspace{-3px}
\begin{gather}
    \mathbf{x}^{d-} = \mathbf{x}^{fc} + B_{P}(\vec{\theta}; \mathcal{P}) + B_{E}(\vec{\psi}; \mathcal{E}) \\
    \mathbf{x}^{d} = \text{LBS}(\mathbf{x}^{d-}, \mathbf{J}(\psi), \theta, \mathcal{W}),
\end{gather}
where $\mathbf{x}^{d-}$ denotes the point after applying the blendshapes and before applying transformation via linear blend skinning (LBS). $\psi$, $\theta$, $\beta$ are the expression, pose, and shape parameters of the FLAME model, respectively, for animating the avatar, and $\mathcal{E}, \mathcal{P},$ and $\mathcal{W}$ are the expression blendshapes, pose blendshapes, and LBS weights, respectively, which are estimated by an MLP. The normal of each point $\mathbf{n}_c$ is defined as a signed distance field (SDF), which is the canonical network's output as follows: $\mathbf{n}_c = \nabla_{\mathbf{x}_{c}}\text{SDF}(\mathbf{x}_{c})$. The normal of the deformation space $\mathbf{n}_d$ is represented by a deformation network which deforms the canonical point set $P_c$ to the deformed point set $P_d = \{\mathbf{x}_d^i\}$. The point deformation is fully differentiable, so it can define the normal deformation as follows:
\vspace{-1px}
\begin{equation}\label{eq:normal_deformation}
    \mathbf{n}_d = l \mathbf{n}_c \left({\partial \mathbf{x}_d \over \partial \mathbf{x}_c} \right)^{-1},
\end{equation}
where $l$ denotes the normalizing factor, which ensures the output of normal value should be the unit length. The RGB of a point is represented by $c_d = s_d \circ a$, the Hadamard Product of the shading $s_d$, and albedo $a$.
\vspace{-2px}

\section{Method} \label{sec:method}
\vspace{-3px}
\subsection{Personalized Generative Avatar Model}\label{sec:model}
\vspace{-3px}
Our generative avatar model takes a latent code $\mathbf{z} \in \mathbb{R}^{(D+1) \times d}$ and FLAME parameters $\beta$, $\theta$, and $\psi$ as inputs. The latent code is the concatenation of the $D$ part-wise latent codes $\{\mathbf{z}^j\}_{j=0...D}$, where each part-wise latent code  $\mathbf{z}^j \in \mathbb{R}^d$ controls the identity of the humans or the subpart such as hair and nose. We treat that $\mathbf{z}^0$ controls the overall identity variations while changing other codes $\mathbf{z}^{j\neq0}$ varies only the subparts of the face, preserving the same identity represented by $\mathbf{z}^0$. By changing FLAME parameters, we can animate the avatars to have varying face poses and expressions. The shape parameter of FLAME $\beta$ also affects the overall coarse shape of the avatar, and we assume the parameter is fixed for the same individual with the same $\mathbf{z}^0$. By extending the PointAvatar~\cite{zheng2023pointavatar}, our avatar model is represented by a set of \emph{generic} (or person-agnostic) canonical point ${P}^{gc} = \{ \mathbf{x}_i^{gc}\}_{i=\{1,\cdots, N\}}$. To this end, our model, $\mathcal{M}_{\phi}$, infers the 3D point locations in the deformed space $\mathbf{x}_i^{d}$, normal vector $\mathbf{n}_i\in \mathbb{R}^3$, and the albedo color $\mathbf{a}_i \in \mathbb{R}^3$ for each queried canonical point as:
\vspace{-1px}
\begin{equation}\label{eq:network_io}
    \mathcal{M}_{\phi}(\mathbf{x}_i^{gc}, \mathbf{z}, \vec{\beta}, \vec{\theta}, \vec{\psi}) = \{ \mathbf{x}_i^{d}, \mathbf{n}^d, \mathbf{a}_i \},
\end{equation}
where $\mathbf{x}_i^{d}$ represents the 3D point after applying identity and appearance variations controlled by $\mathbf{z}$, as well as the facial pose and expression changes by FLAME parameters. \figref{fig:overview} represents an overview of PEGASUS.

In contrast to the original PointAvatar, which represents a single identity only, we train a single avatar model to represent multiple face appearances, where appearance can vary by changing disentangled latent codes $\mathbf{z}$. We tackle this challenging problem by introducing the generic canonical space, which is person-agnostic. 
While our model ideally expresses a range of identities, we observe that representing extremely diverse individuals with a single implicit model often results in blurry avatars, as demonstrated in our ablation studies.
Yet, we demonstrate that our model can successfully achieve the goal of a personalized generative avatar model, allowing face part variation while preserving the same identity. Importantly, in order to build the personalized generative avatar model, we present a way to synthesize the dataset of the target individual via part-swapping, described in \cref{sec:synthetic}.

\noindent \textbf{Multi-stage Canonical Spaces and Point Deformation.} 
While the original PointAvatar considers two-staged deformation (canonical, FLAME-canonical, and deformed space), we consider one more stage, resulting in generic canonical ($gc$), subject-specific canonical ($sc$), subject-specific FLAME-canonical ($fc$), and deformed space ($d$). We empirically find that introducing the generic canonical space enables us to avoid bad local minima in training the model with multiple face appearances, which is similar to PointAvatar, and to enhance identity preservation while varying the attributes by randomly sampled latent code. See more details in Supp. Mat.

The generic canonical space and the point locations defined in this space ${P}^{gc} = \{ \mathbf{x}^{gc}_i\}_{i=1...N}$ are shared among all identities.
We first map the points $\mathbf{x}^{gc}_i$ from the generic canonical space into the subject-specific canonical space ${P}^{sc} = \{ \mathbf{x}^{sc}_i\}_{i=1...N}$ by adding point offsets $\mathcal{O}_i^{gc\rightarrow sc}$ that are conditioned by latent code $\mathbf{z}$. Subsequently, we map the points in the subject-specific canonical space into the FLAME-canonical space via another point offset $\mathcal{O}_i^{sc\rightarrow fc}$, similar to the PointAvatars. That is,
\vspace{-1px}
\begin{align}\label{eq:cc_to_sc}
    \mathbf{x}^{sc}_i = \mathbf{x}^{gc}_i + \mathcal{O}_i^{gc\rightarrow sc} \\
    \mathbf{x}^{fc}_i = \mathbf{x}^{sc}_i + \mathcal{O}_i^{sc\rightarrow fc},
\end{align}
where $\mathcal{O}_i^{gc\rightarrow sc}$ and $\mathcal{O}_i^{sc\rightarrow fc}$ are inferred from the learned deformation MLP model. Intuitively, our subject-specific canonical space is equivalent to the ``canonical space'' of PointAvatars, where we introduced one more prior stage to handle multiple identities.  

As in PointAvatars, we use a coordinate-based MLP to infer deformation offsets, blendshapes, and LBS weights:
\vspace{-2px}
\begin{equation}\label{eq:deformation_decoder_io}
    MLP(\mathbf{z}, \mathbf{x}_i^{gc}) = \{ \mathcal{O}_i^{gc\rightarrow sc},  \mathcal{O}_i^{sc\rightarrow fc}, \mathcal{E}, \mathcal{P}, \mathcal{W} \}.
\end{equation}
The deformed point is then computed as:
\vspace{-2px}
\begin{gather}
\label{eq:fc_to_bs}
    \mathbf{x}^{d-} = \mathbf{x}^{fc} + B_{S}(\vec{\beta}; \mathcal{S}) + B_{P}(\vec{\theta}; \mathcal{P}) + B_{E}(\vec{\psi}; \mathcal{E})\\
    \mathbf{x}^{d} = \text{LBS}(\mathbf{x}^{d-}, \mathbf{J}(\psi), \theta, \mathcal{W}).
\end{gather}
Different from PointAvatar, we leverage the shape blendshapes basis $B_{S}$ of the FLAME, allowing us to change the coarse shape of the avatar by controlling the shape parameter $\beta$ of the FLAME, which is useful for building our synthetic DB to enable better face alignments described in \cref{sec:synthetic}.

\begin{figure}[t]
\includegraphics[trim={0 0 0 0},clip,width=1.0\columnwidth]{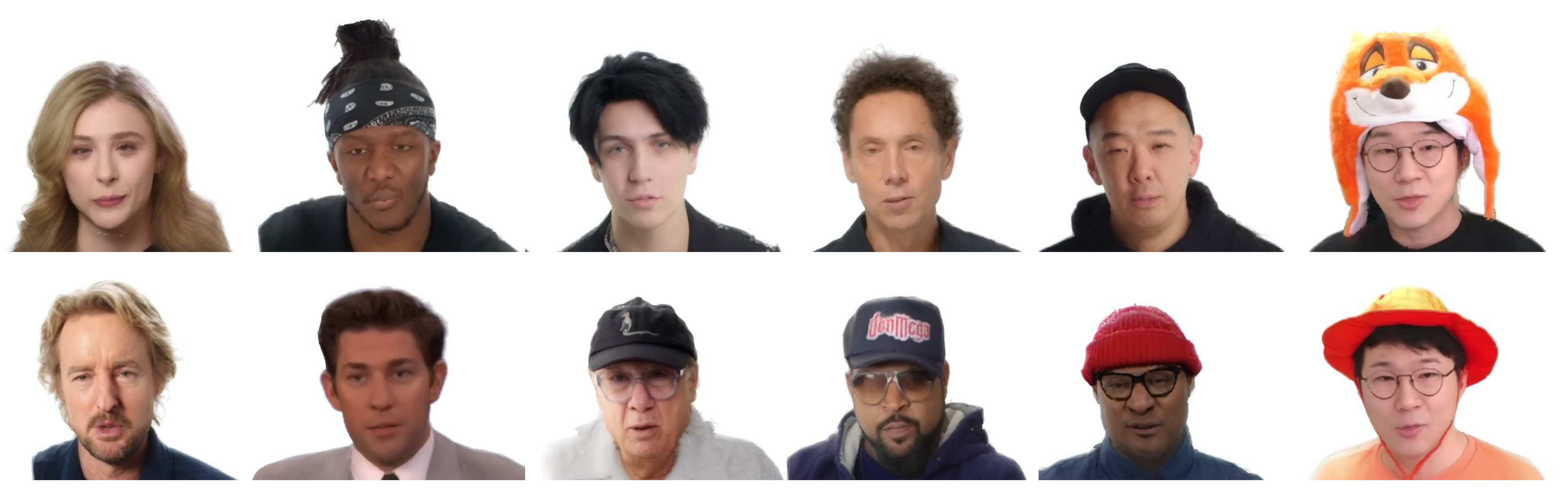}
\vspace{-5px}
\caption{\textbf{DB Avatar.} We create deformable avatar models from the attribute DB videos, which are monocular RGB inputs. We show some examples of our collection of avatars from attribute DB videos with the same FLAME parameters.}
\label{fig:db_videos}
\end{figure}
\begin{figure}
\centering
     \begin{subfigure}[t]{0.14\columnwidth}
         \centering
         \includegraphics[width=\columnwidth]{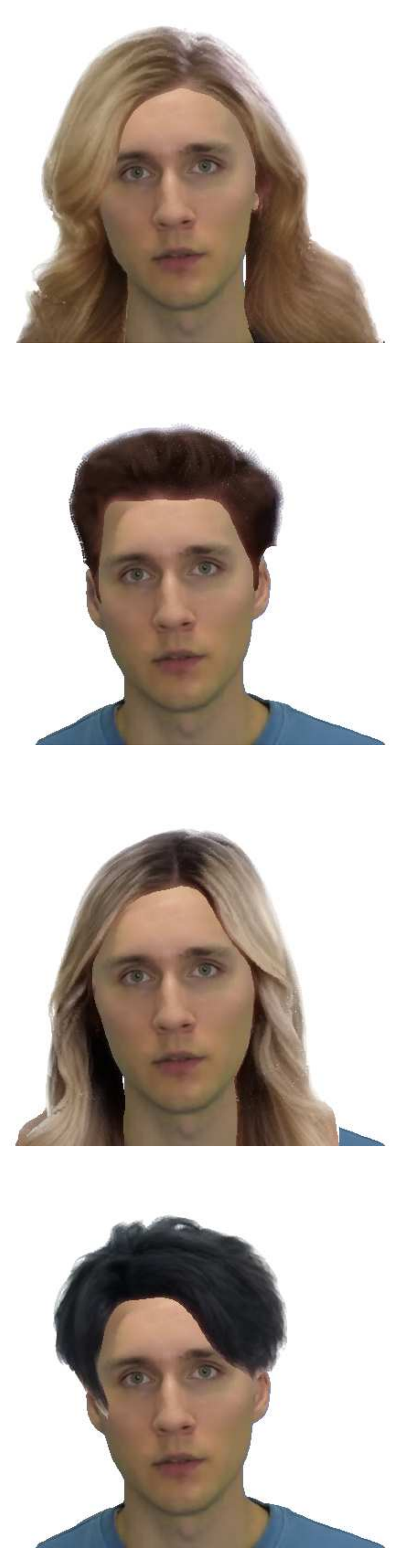}
         \caption{Hair}
         \label{fig:hair}
     \end{subfigure}
     \begin{subfigure}[t]{0.135\columnwidth}
         \centering
         \includegraphics[width=\columnwidth]{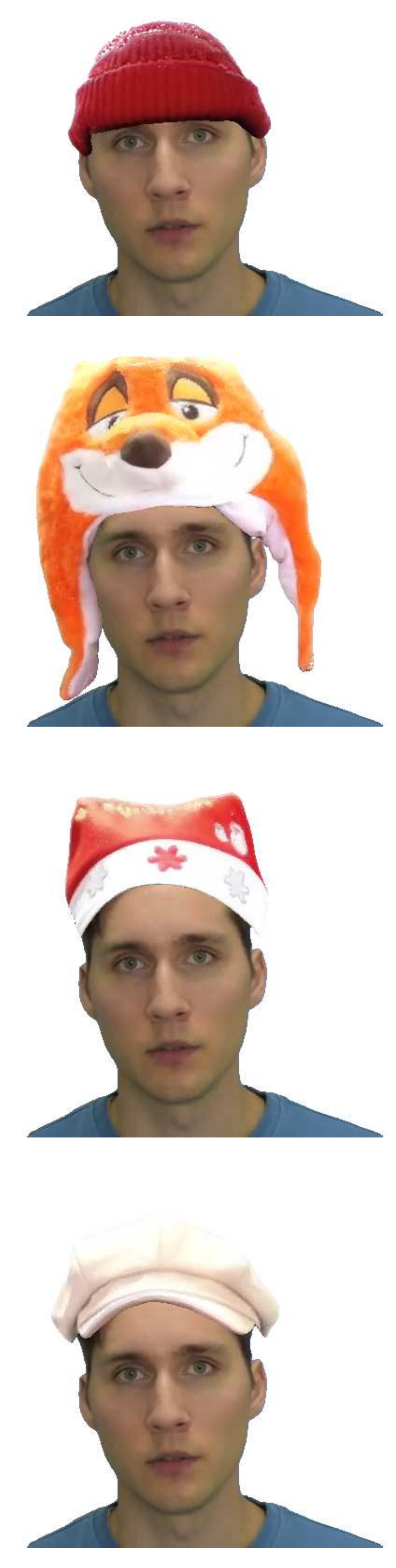}
         \caption{Hat}
         \label{fig:hat}
     \end{subfigure}
     \begin{subfigure}[t]{0.137\columnwidth}
         \centering
         \includegraphics[width=\columnwidth]{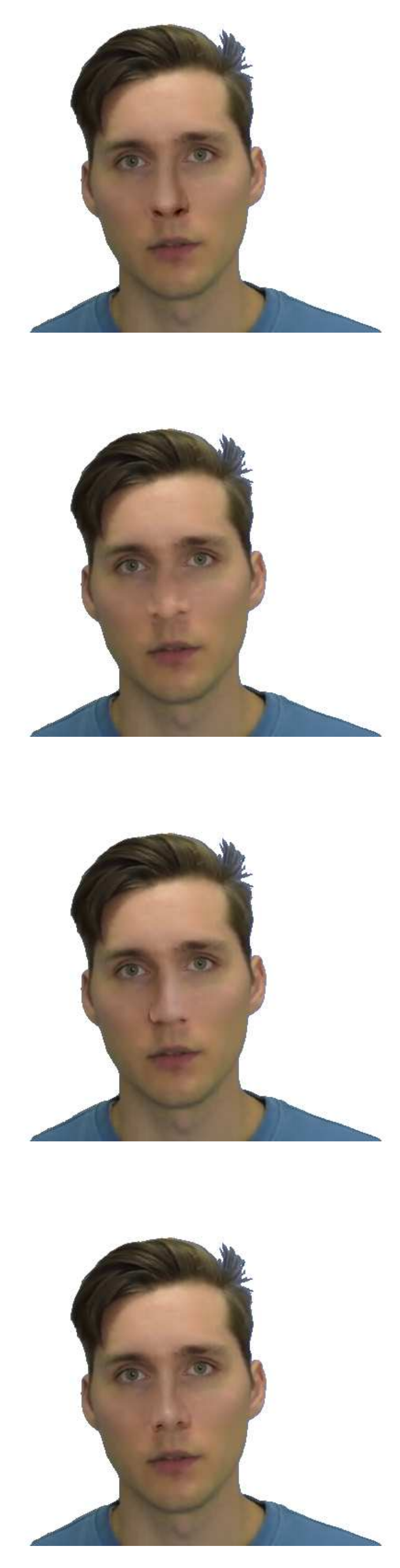}
         \caption{Nose}
         \label{fig:nose}
     \end{subfigure}
     \begin{subfigure}[t]{0.134\columnwidth}
         \centering
         \includegraphics[width=\columnwidth]{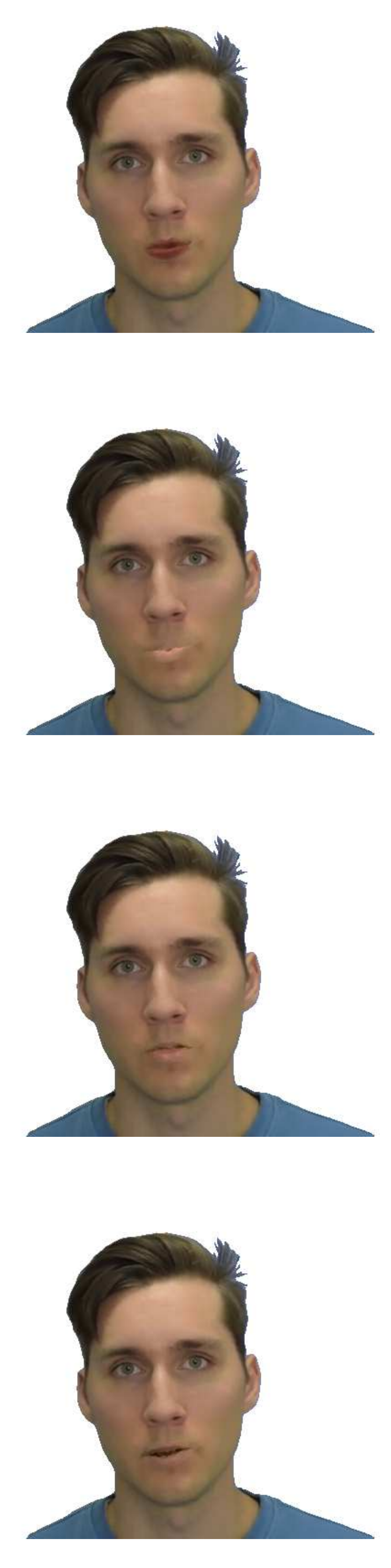}
         \caption{Mouth}
         \label{fig:mouth}
     \end{subfigure}
     \begin{subfigure}[t]{0.141\columnwidth}
         \centering
         \includegraphics[width=\columnwidth]{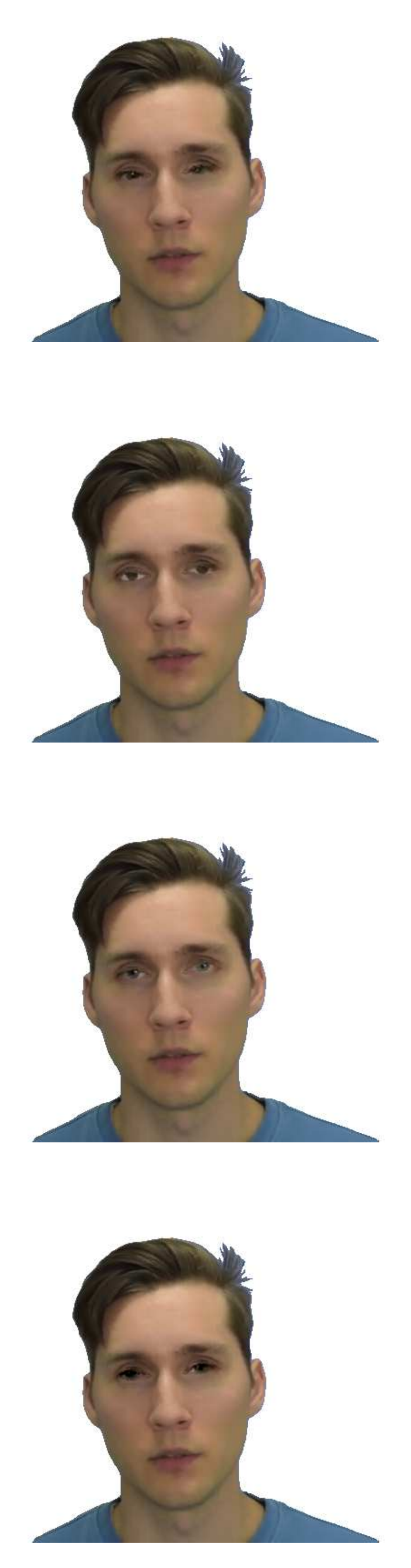}
         \caption{Eyes}
         \label{fig:eyes}
     \end{subfigure}
     \begin{subfigure}[t]{0.130\columnwidth}
         \centering
         \includegraphics[width=\columnwidth]{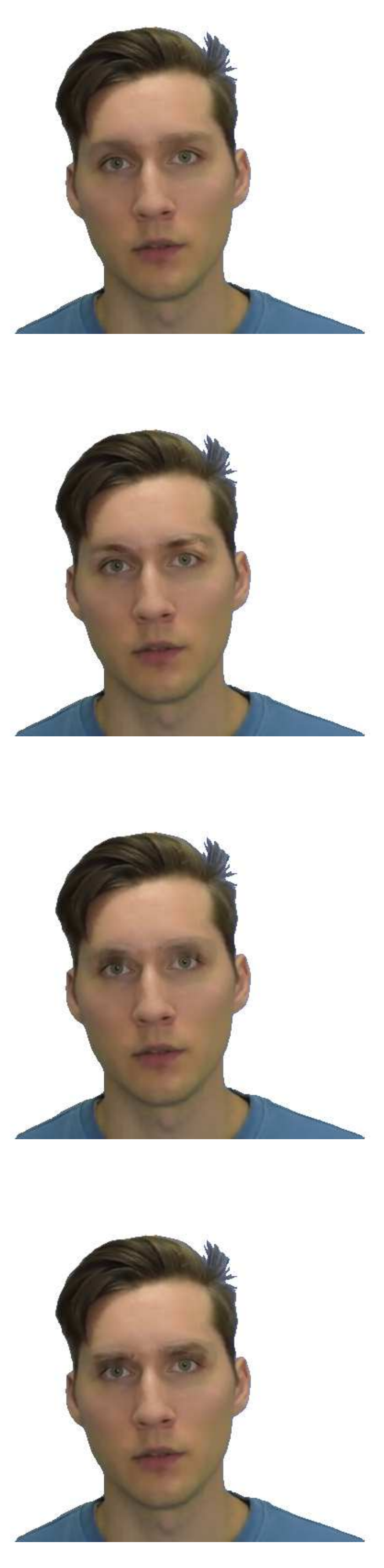}
         \caption{Brows}
         \label{fig:eyebrows}
     \end{subfigure}
     \vspace{-5px}
     \caption{\textbf{Part-Swapped Videos of the Target Individual.} Some examples of the synthetic DB created through part-swapping. Our synthetic DB includes a variety of hair, hats, eyes, noses, mouths, and eyebrows.}
\label{fig:synthesized_videos}
\end{figure}
\noindent \textbf{Canonical Representations.}
We utilize an MLP to infer the SDF value $\sigma_i  \in \mathbb{R}$, albedo $\mathbf{a}_i \in \mathbb{R}^3$, shading $\mathbf{s}_i \in \mathbb{R}^3$, point segment cues $\chi_i \in [0, 1]$ for the $i$-th point at the subject canonical space $\mathbf{x}_i^{sc}$:
\vspace{-1px}
\begin{equation}\label{eq:deformation_decoder_io}
    MLP(\mathbf{z}, \mathbf{x}_i^{sc}) = \{ \sigma_i, \mathbf{a}_i, \chi_i \}.
\end{equation}
Note that we consider these cues on the subject canonical space $\mathbf{x}_i^{sc}$, rather than the generic canonical space since we empirically find inferring it in the generic canonical space suffers from local minima issue. Similar to PointAvatar, the SDF cues are utilized to infer surface normal in the subject canonical space $\mathbf{n}_{sc}$ and the ones in the deformed space can be computed as in \cref{eq:normal_deformation}. Note that, different from PointAvatar, the cues in the canonical space are also conditioned by latent code $\mathbf{z}$, allowing for varying by controlling latent codes for part appearance changes. Furthermore, we additionally include the binary segmentation cues $\chi_i$ to estimate a ``synthesis'' part in the current identity represented by $\mathbf{z}$, which is used in our \emph{Zero Shot Transfer} approach in \secref{sec:zeroshot}.

\noindent \textbf{Comparison over PointAvatar.}
The major difference from the PointAvatar is the use of latent codes $\mathbf{z}$ to enable the single model can handle varying appearance changes. For this purpose, we modify the model, including a generic canonical stage and injecting the $\mathbf{z}$ into the submodules. We also made several modifications, including (1) the beta controlling part in \eqnref{eq:fc_to_bs}, which is essential for fitting the subject and face attribute when generating a synthetic database, (2) inferring the segmentation mask for the usage in \emph{Zero Shot Transfer}.
\vspace{-2px}

\subsection{Synthetic DB Generation via Part Swapping}
\label{sec:synthetic}
\vspace{-2px}
We aim to build our personalized generative model to preserve the target human identity, while allowing changing facial attributes, such as hair, nose, or wearing a hat. To learn such a model, we need the videos of the target human with all such variations, which is not available in practice. We present a solution to synthesize such variations from other video sources by swapping a face subpart of the target identity person with others. Examples are shown in \figref{fig:synthesized_videos}. We collect a set of monocular videos, denoted as $V^{db} = \{V_i, p_i\}_{i=1\cdots K}$ from various individuals to model various types of facial attribute variations. For each video $V_i$, we determine the target facial attribute $p_i \in \mathbf{P}$ which we want to use for the swapping, where $\mathbf{P} = \{\text{hair, nose, hat, eyes, eyebrows, mouth}\}$.

For each monocular video $V_i$ from the facial attribute DB, we build a personalized avatar $\mathcal{M}^{db}_i$ using our modified avatar generation module which removes the subject-canonical space with only the single video identity as shown in \figref{fig:db_videos}. Here we set the identity latent code $\mathbf{z}_0$ as learnable while setting other parts accordingly. Once built, the avatar is animatable following the FLAME parameters. 

\noindent \textbf{Face Part Swapping.}
We denote the input video of the target person as $V^{tp}$. The goal of our face part swapping is to replace the facial attributes  $V^{tp}$ with the one using the person appeared in by $i$-th attribute video $V_{i}$. Since both videos have different poses, viewpoints, and facial expressions of different individuals, such replacement is non-trivial in 2D video space. Our idea is to leverage the animatable avatar model $\mathcal{M}^{db}_{i}$ constructed from $V_{i}$ to render the facial attribute aligned into the target identity's videos, $V^{tp}$. This can be performed by inputting the FLAME parameters and camera parameters obtained from $V^{tp}$ into $\mathcal{M}^{db}_{i}$ and by rendering only the necessary attribute region with blending. To choose the selected attribute regions specified by the corresponding attribute $p_i \in \mathbf{P}$, we use an off-the-shelf face part segmentation model~\cite{yu2018bisenet} to obtain the mask of the desired target attribute $p_i$. 

Then, we can synthesize the attribute part into the target human videos $V^{tp}$ as follows:
\vspace{-5px}
\begin{gather}
    I_{\text{i-th attributes}} = \mathcal{R}(\mathcal{M}^{db}_{i}(\vec{\theta}_{\text{tp}}, \vec{\beta}_{\text{tp}}, \vec{\psi}_{\text{tp}})) \nonumber \\
    I_{\text{target-swapped}} = \boldsymbol{1}_{\text{tp}} \cdot I_{\text{tp}} + \boldsymbol{1}_{\text{i-th attributes}} \cdot \mathcal{B}(I_{\text{i-th attributes}}),
    \nonumber
\end{gather}
where $\mathcal{R}$ denotes the rendering function from the avatar $\mathcal{M}^{db}_i$ with the FLAME parameters obtained from $V^{tp}$. $\boldsymbol{1}_{\text{tp}}$ is the segmentation mask to select the target subject regions excluding the attribute parts, and $\boldsymbol{1}_{\text{i-th attributes}}$ is segmentation for the attributed part of from $\mathcal{M}^{db}_i$'s rendering, respectively. Note that we use the shape parameter of the target human $\beta_{\text{subject}}$ on $\mathcal{M}^{db}_i$ as an input to make better alignment into the target identity, which was the motivation for introducing the shape parameter $\beta$ in building our avatar, different from the original PointAvatar model. $\mathcal{B}$ denotes the blending function, where we use Poisson Blending~\cite{perez2023poisson} to reduce artifacts. We further perform post-processing to enhance the quality of part-swapped images using OpenCV's dilate and erode function to remove holes. As a special preprocessing for hair-swapping, it is empirically advantageous to synthesize the target person's hair into a bald head before the blending, where we leverage Stable Diffusion~\cite{rombach2022high} with auto-generated mask images. See Supp. Mat. for details. 

We denote $\hat{V}^{tp}_i$ as the part-swapped videos by $i$-th attribute DB identity. Examples are shown in ~\figref{fig:synthesized_videos}. Note that the resulting videos contain the same target identity with varying attributes via synthesis, which we use to build our personalized generative models. 

\subsection{Learning for Personalized Generative Model}\label{sec:loss}
\vspace{-2px}
\begin{figure}[t]
\includegraphics[trim={0 0 0 0},clip,width=\columnwidth]{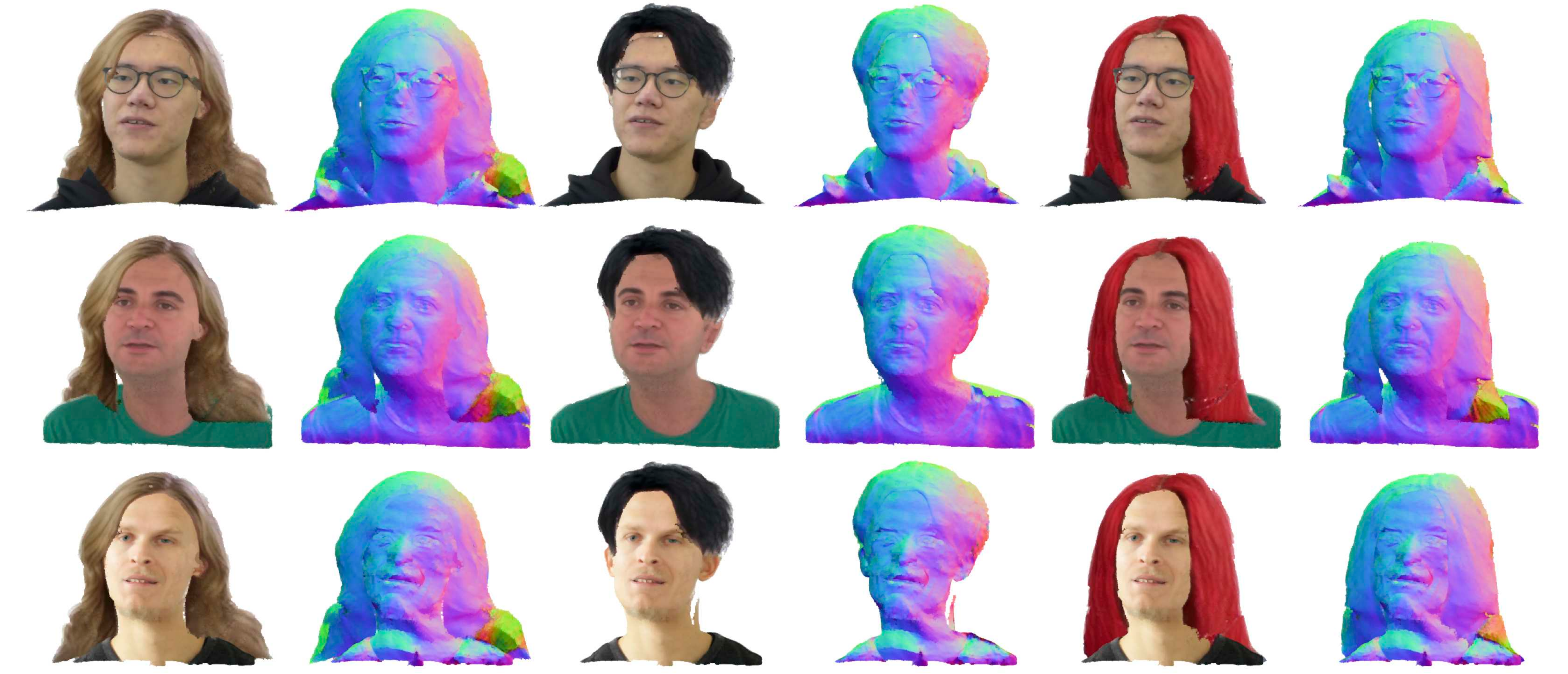}
\caption{\textbf{Zero Shot Transfer.} PEGASUS generates high-quality and natural appearances through zero-shot transfer.}
\label{fig:zero_shot_transfer}
\vspace{-10px}
\end{figure}
\noindent \textbf{Latent Code Setting.}
We train our model by using $V^{tp}$ and synthesized videos $\{ \hat{V}^{tp}_i\}_{ i=1\cdots K  }$. For each video, we set the latent code $\mathbf{z} = \{\mathbf{z}^p\}_{p=0, \cdots, D}$ according to the attribute types. Specifically, we use the same shared learnable identity latent code $\mathbf{z}_0$ for all videos, given that the videos are for the same identity. If a video is about the variation of $p$-th attribute category, where $p \in \mathbf{P}$, we assign a separate learnable latent code for that part $\mathbf{z}^p$, by keeping other latent code parts shared. With this setup, we allow the model can have the latent codes in a disentangled manner so that each attribute code part can represent the corresponding facial subparts.  

\noindent \textbf{Loss Function.}
We follow the PointAvatar~\cite{zheng2023pointavatar} to define loss functions. The total loss is as follows:
\vspace{-1px}
\begin{equation}\label{eq:total_loss}
\begin{split}
    \mathcal{L} & = \lambda_{\text{rgb}}\mathcal{L}_{\text{rgb}} + \lambda_{\text{mask}}\mathcal{L}_{\text{mask}} + \lambda_{\text{FLAME}}\mathcal{L}_{\text{FLAME}} + \lambda_{\text{vgg}}\mathcal{L}_{\text{vgg}}\\
    & + \lambda_{\text{normal}}\mathcal{L}_{\text{normal}} + \lambda_{\text{seg}}\mathcal{L}_{\text{seg}} + \lambda_{\mathbf{z}~\text{reg}}\mathcal{L}_{\mathbf{z}~\text{reg}},
    \nonumber
\end{split}
\end{equation}
where $\mathcal{L}_{\text{rgb}}$, $\mathcal{L}_{\text{mask}}$, $\mathcal{L}_{\text{FLAME}}$ penalize RGB, mask, FLAME parameter differences respectively. $\mathcal{L}_{\text{vgg}}$ is based on the VGG feature to enhance the rendered image quality. Different from previous work~\cite{zheng2023pointavatar, zheng2022imavatar}, we also include three more losses, $\mathcal{L}_{\text{normal}}$, $\mathcal{L}_{\text{seg}}$ and $\mathcal{L}_{\mathbf{z}~\text{reg}}$. We adopt the normal loss as follows: $ \mathcal{L}_{\text{normal}} = \|\mathbf{n} - \mathbf{n}^d\|$, and we empirically find its advantage in producing better-quality avatars. We generate the pseudo ground truth normal $\mathbf{n}$ from the $V^{tp}$ and the avatar trained with a single identity of each $V^{db}$.
We also include segmentation loss to predict facial attribute categories $\chi_i$ per each point. See more details in Supp. Mat.

\noindent \textbf{Training Strategy.}
We train PEGASUS in a coarse-to-fine manner. First, following PointAvatar, we upsample the number of points and reduce the radii of the points during the training with the constant period of epochs. Second, we train our model in the two-step strategy. Initially, we train our model using the target individual $V^{tp}$, which is no part swapped on the face, with the latent codes $\mathbf{z}$ until the beginning of the training. Subsequently, we use all of the part-swapped videos $\hat{V}_i^{tp}$ until the end of training. Check the details in Supp. Mat.
\vspace{-2px}

\section{Generative Avatar via Zero-Shot Transfer} \label{sec:zeroshot}
\vspace{-2px}
We present an alternative method to efficiently achieve the goal of a personalized generative avatar without producing part-swapped synthesized videos. Our core idea is based on the assumption that we already have the previously constructed personalized avatar model for an identity (denoted as the \emph{source human}), $\mathcal{M}_{\phi}$, with the functionally to control the face attribute variations. Given the new identity's video (denoted as the \emph{target human}), we first train our generative avatar architecture with the single video of the target human, resulting in $\mathcal{M}_{th}$. Then, we aim to achieve the same goal of the personalized avatar for the target human, by fusing the controlled attributed part of  $\mathcal{M}_{\phi}$ and the remaining part $\mathcal{M}_{th}$, which we call a ``zero-shot model''. Especially, given the FLAME parameters and input latent codes inputs, we can drive both models as:
\vspace{-5px}
\begin{align}\label{eq:zero_shot_sh}
    \mathcal{M}_{\phi}(\mathbf{x}^{gc}, \mathbf{z}, \vec{\beta}, \vec{\theta}, \vec{\psi}) = \{ \mathbf{x}_{\phi}^{d}, \mathbf{n}_{\phi}^{d}, \mathbf{a}_{\phi}, \chi_{\phi} \}, \\
    \mathcal{M}_{th}(\mathbf{x}^{gc}, \mathbf{z}, \vec{\beta}, \vec{\theta}, \vec{\psi}) = \{ \mathbf{x}_{th}^{d}, \mathbf{n}_{th}^{d}, \mathbf{a}_{th}, \chi_{th} \}.
\end{align}

The final version of the avatar is constructed by combining the subsets of point clouds from both avatars, using the estimated segmentation masks, $\chi_{\phi}$ and $\chi_{th}$:
\vspace{-5px}
\begin{align}\label{eq:zero_shot_naive_pcd}
    P_{\text{naive}} = \{ \mathbf{x}_{\phi}^{d, i} \}_{\chi^{i}_{\phi} = 1} \cup  \{ \mathbf{x}_{th}^{d, i} \}_{\chi^{i}_{th} = 0},
\end{align}
where $\chi^i_{\phi}$ and $\chi^i_{th}$ are the segmentation masks of the face attribute we currently try to control via $\mathbf{z}$. Intuitively, we replace the points of the desired facial attribute of the novel target human with those of the pretrained generative avatar. While we find this naive composition is already compelling, we observe that there exists a gap between the fused parts. To enhance the quality, we further perform an additional optimization processing to better alignment, with color blending. See Supp. Mat. for the post-processing. Examples of our zero-shot modeling are shown in Fig.~\ref{fig:zero_shot_transfer}.
\vspace{-2px}

\section{Experiments} \label{sec:experiments}
\vspace{-2px}
\begin{figure}[t]
\includegraphics[trim={0 0 0 0},clip,width=1.0\columnwidth]{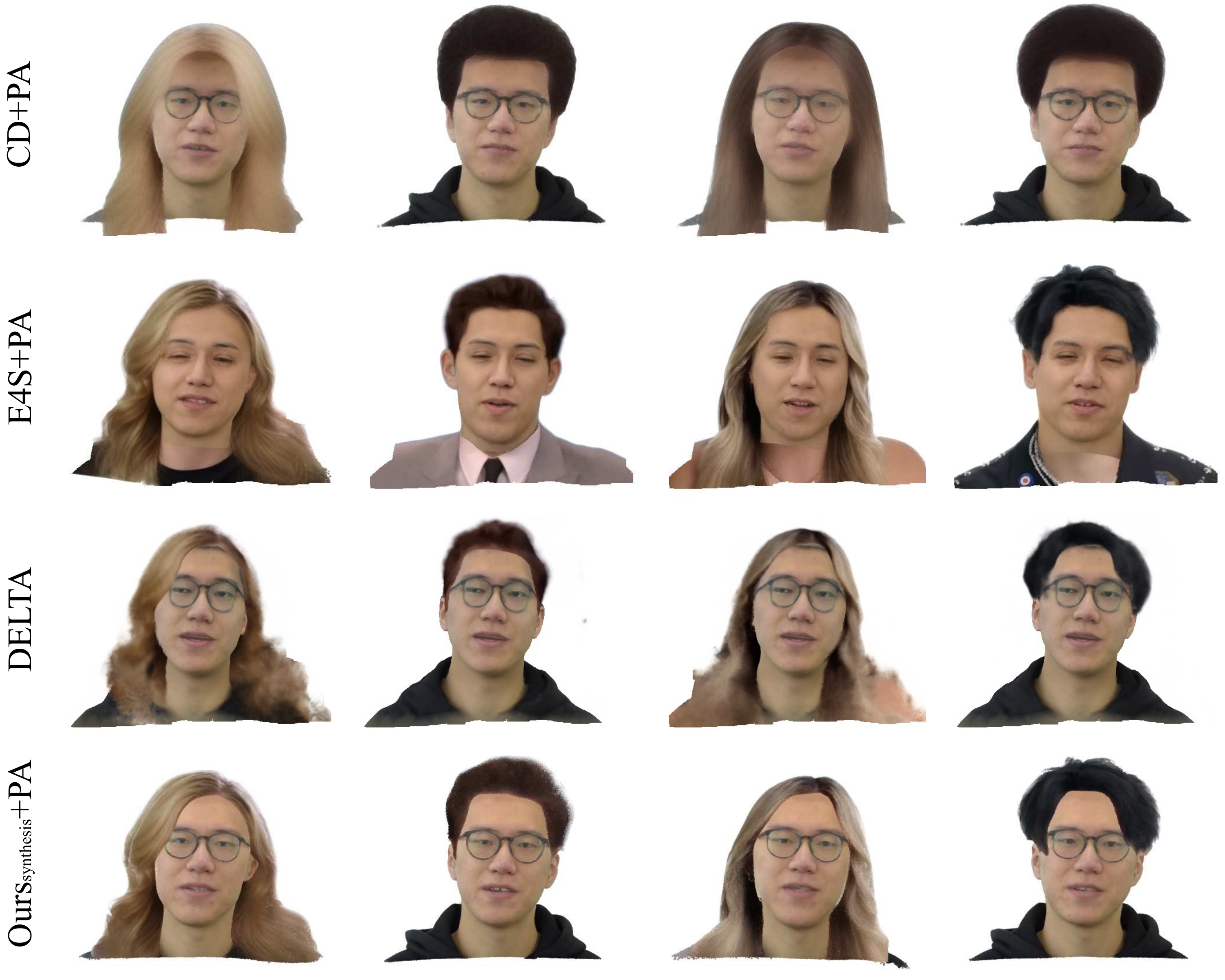}
\vspace{-15px}
\caption{\textbf{Single Part-Swap Avatar on Hair.} Our synthesis method creates a photo-realistic avatar with a hairstyle that is accurately transferred.}
\label{fig:qualitative_results_baselines}
\vspace{-10px}
\end{figure}
\noindent \textbf{Datasets.}
As the attribute database, we collect publicly available 109 videos from the Internet, and build their individual DB avatar model $\mathcal{M}^{db}$ as shown in \figref{fig:db_videos}. For the target person $V^{tp}$ used of the personalized generative avatar, we select the publicly available videos from NerFACE~\cite{gafni2021dynamic}, and the individuals are shown in \figref{fig:synthesized_videos} and \figref{fig:qualitative_results_baselines}. To reenact the reconstructed avatar into unseen facial poses and expressions, we extract FLAME parameters using DECA~\cite{feng2021learning} from our own monocular video with diverse facial orientations and expressions.

\subsection{Part-Swapping Comparison with Baselines}
\label{subsec:eval-part-swap-avatar}
\vspace{-2px}
Given that we are the first to build a personalized generative model, there is no direct competitor to compare the full generative functionality. Thus, we consider a sub-problem of building an animatable 3D avatar by transferring a facial attitude from another video source. While the resulting output is not a generative model due to its limitation of producing unseen attributes, one can use this strategy to alter parts of the face, assuming a large number of attribute source videos are available. In this evaluation, we only consider hairstyles as our attribute and consider 5 videos with different hairstyles. Examples are shown in \figref{fig:qualitative_results_baselines}.

\noindent \textbf{Baselines.} 
We consider possible alternative approaches to building the 3D avatar of the target individual with the hair from another video. 

\emph{DELTA~\cite{feng2023learning}:} DELTA achieves the transfer of hairstyles from a source to a target by employing a hybrid approach that combines both explicit and implicit representations. The major goal of DELTA is aligned with this sub-problem test, while it does not have generative functionality.

\emph{E4S~\cite{liu2023fine} + PointAvatar (E4S+PA):} E4S employs GAN inversion for the face swapping. As a way of building a 3D avatar, we first replace the hair of the target individual in 2D spaces on all image frames via E4S. Then, we apply the original version of PointAvatar to make it into a 3D avatar model. Note that the GAN-based method does not guarantee the view consistency on the synthesized images, resulting in blurry 3D avatar construction. 

\emph{Custom Diffusion~\cite{kumari2023multi} + PointAvatar (CD+PA):} Similar to the E4S+PA, we can apply the Custom Diffusion model as a tool to produce hair-changed 2D images for the target individual, conditioned by the hair-style of other video source. Then, we apply the original PointAvatar. 

\emph{$\text{Ours}_\text{swap}$ + PointAvatar (Ours$_\text{swap}$ + PA):} We also include a simplified version of ours as a baseline, where we produce the part-swapped 2D videos (described in Sec.~\ref{sec:synthetic}) for each hairstyle transfer and apply PointAvatar. 

\emph{$\text{Ours}_\text{person-gen}$ and $\text{Ours}_\text{zero-shot}$ :} We show the performance of our generative models using the latent codes corresponding to the target hairstyles. Note our models can produce not just various hairs, but all other attribute styles.

\noindent \textbf{Metrics.}
After we build 3D avatars of the target individual by transferring the hairstyle from video sources, we apply unseen facial expressions and head orientation to visualize the avatar in diverse novel poses and render them into images. For the comparison, we consider both the naturalness or the 3D avatar and identity preservation of the target individual. We use two metrics, Fréchet Inception Distance (FID) and Kernel Inception Distance (KID), to evaluate the naturalness of the rendering of the produced 3D avatar. In computing FID and KID, we compare the distributions of rendered outputs of the 3D avatars with the background matted FFHQ~\cite{karras2019style}. To quantify whether the output 3D avatars preserve the original identity of the target human, we include ArcFace~\cite{deng2019arcface} metric. Here, we compare the rendering of the edited version with the rendering of the non-edit PointAvatar with the same unseen face pose. 
\begin{table}[t]
\centering
\small{
\begin{tabular}{l|cc|c}%
\toprule
\multirow{2.5}{*}{Method}
    &\multicolumn{2}{c|}{Naturalness}&\multicolumn{1}{c}{Identity} \\
    \cmidrule(lr){2-3}\cmidrule(lr){4-4} & \thead{FID${\downarrow}$} & \thead{KID${\downarrow}$} & \thead{ArcFace${\uparrow}$} \\ 
\midrule
CD + PA  & 181.60 & \textbf{0.1367} & 0.6691 \\
E4S + PA & 176.64 & 0.1416 & 0.5701 \\
DELTA & 198.40 & 0.1797 & 0.6732 \\
Ours$_\text{swap}$+PA & \textbf{169.54} & 0.1406 & \textbf{0.7179} \\
\midrule
Ours$_\text{person-gen}$ & \textbf{190.10} & \textbf{0.1696} & 0.6883 \\
Ours$_\text{zero shot}$ & 191.47 & 0.1881 & \textbf{0.7792} \\
\bottomrule
\end{tabular}
}
\vspace{-5px}
\caption{\textbf{Quantitative Comparison.} The synthesis method (upper rows) and full model of hair category (lower rows).}
\vspace{-5px}
\label{tab:quantiative_hair}
\end{table}

\noindent \textbf{Results.}
We show the quantitative comparison in \tabref{tab:quantiative_hair} and example results in \figref{fig:qualitative_results_baselines}. As shown in the table, the 3D avatar produced by our face-swap \emph{Ours$_\text{swap}$+PA} achieves the best metrics at FID and ArcFace, showing better naturalness while keeping the identity of the target individual. Although the custom diffusion-based output \emph{CD+PA} shows the better result in the KID metric, it changes the identity significantly, resulting in low performance in the ArcFace metric. Our full generative model \emph{Ours$_\text{person-gen}$} also shows convincing performance even though the model is much more generic and trained to express diverse variations. It outperforms all other baseline methods in preserving identity while showing comparable naturalness. Our zero-shot generative model \emph{Ours$_\text{zero-shot}$} shows the best identity-preserving performance because its face part is identical to the non-edited PA while transferring the hair part from \emph{Ours$_\text{person-gen}$}. 

\subsection{Evaluating Generative Performance}
\vspace{-2px}
\begin{table}[t]
\centering
\small{
\begin{tabular}{l|cc|c}%
\toprule
\multirow{2.5}{*}{Method}
    &\multicolumn{2}{c|}{Naturalness}&\multicolumn{1}{c}{Identity} \\
    \cmidrule(lr){2-3}\cmidrule(lr){4-4} & \thead{FID${\downarrow}$} & \thead{KID${\downarrow}$} & \thead{ArcFace${\uparrow}$} \\ 
\midrule
Ours$_\text{no synthesis, latent swap}$ & 231.62 & 0.2630 & 0.6285 \\
Ours$_\text{no synthesis, latent interp.}$ & 240.17 & 0.2482  & 0.4722 \\
Ours$_\text{synthesis, latent interp.}$ & \textbf{206.87} & \textbf{0.1839} & \textbf{0.8127} \\
\bottomrule
\end{tabular}
}
\vspace{-5px}
\caption{\textbf{Evaluating Generative Performance.} Quantitative comparison by producing appearance via latent code interpolation.}
\vspace{-5px}
\label{tab:generative-code-iterp-quant}
\end{table}

\begin{figure}[t]
\includegraphics[trim={0 0 0 0},clip,width=\columnwidth]{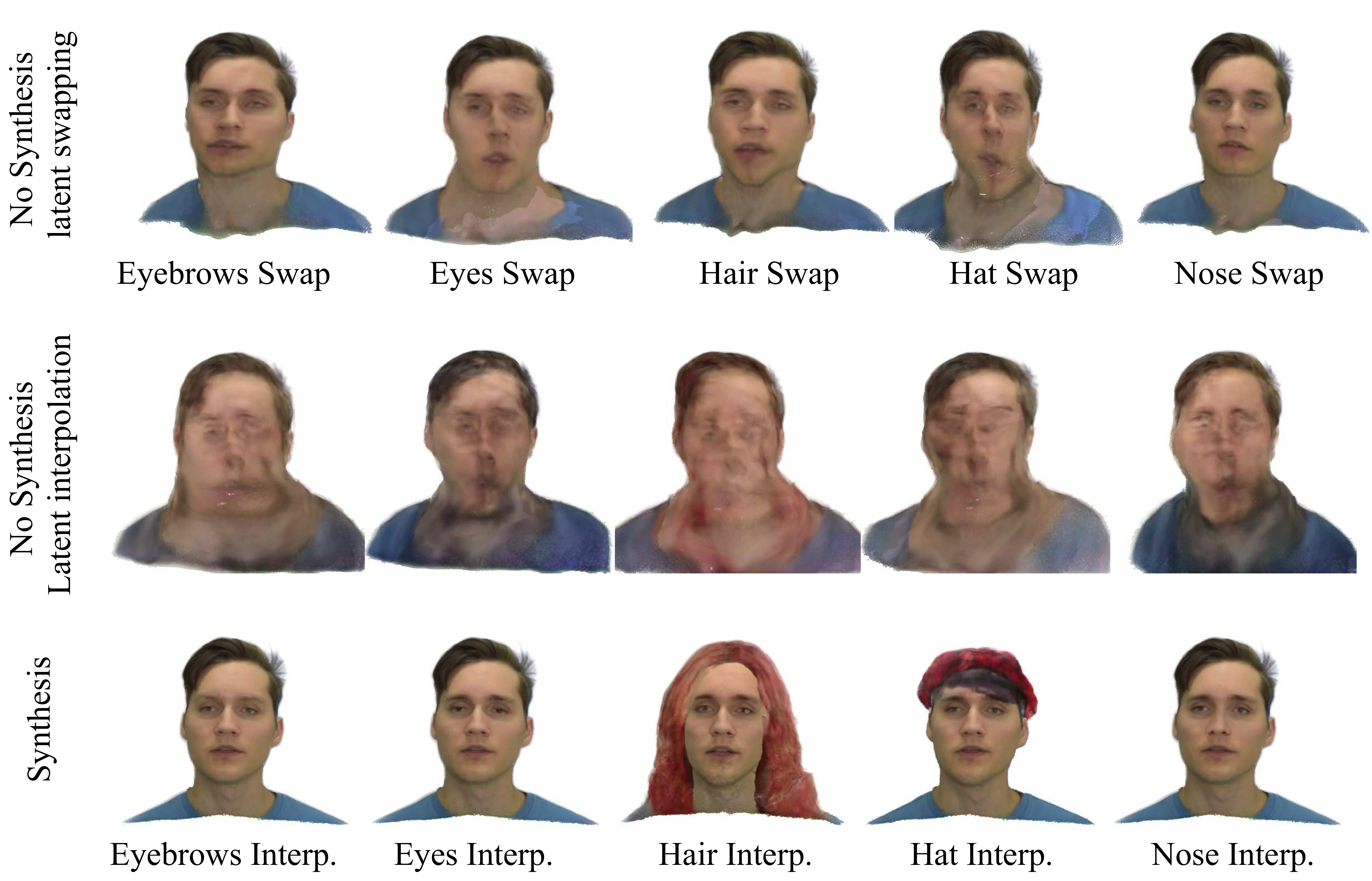}
\vspace{-15px}
\caption{\textbf{Qualitative Comparison.} We compare the generative performance of PEGASUS. Our latent code configuration and synthetic DB generation show compelling generative performance in latent interpolation and swapping compared to other baselines.}
\label{fig:generative-code-iterp-qual}
\vspace{-5px}
\end{figure}

We also compare the generative performance of our models. As the baseline, we consider the scenario of using entire videos, including target individual $V^{tp}$ and face attribute videos $V^{db}$ into the generative model without our facial part-swap approach. Once trained, we check the unseen appearances by interpolating the latent codes of two seen samples during training. However, we consider two ways of interpolation: (1) naive interpolation between the latent codes $(\mathbf{z}_A, \mathbf{z}_B)$ of two original videos (latent interpolation), (2) via latent code swapping by keeping the target individual's latent code $\{ z_p \}_{p\neq i}$ and other sources' attribute latent code $\{ z_q\}_{q=i} $ (latent swapping), $\mathbf{z} = \{z_p, z_q\}_{p \neq i, q=i, i\in [0, n(\mathbf{P})]}$. For quantitative evaluation, we use the same metric as \secref{subsec:eval-part-swap-avatar} to measure the naturalness and identity preserving. The quantitative result is shown in \tabref{tab:generative-code-iterp-quant}, and example qualitative results are shown in \figref{fig:generative-code-iterp-qual}. The outputs of our model show compelling performance in producing realistic face part variations while keeping the identity. As expected, both interpolation strategies of the baseline models struggle to generate realistic avatars for the interpolated latent codes.
\vspace{-2px}

\subsection{Ablation Studies and More Results}
\begin{table}[t]
\centering
\small{
\begin{tabular}{lc|cccc}%
\toprule
$\mathcal{O}_{gc}$ & $\mathcal{O}_{sc}$ & \thead{PSNR$\uparrow$} & \thead{SSIM${\uparrow}$} & \thead{LPIPS${\downarrow}$} \\
\midrule
$\times$ & $\times$ & 20.92 & 0.9059 & 0.1351 \\
$\times$ & $\checkmark$ & 21.55 & 0.9033 & 0.1292\\
$\checkmark$ & $\checkmark$ & \textbf{21.75} & \textbf{0.9059} & \textbf{0.1291} \\
\bottomrule
\end{tabular}
}
\vspace{-5px}
\caption{\textbf{Ablation Study: offsets.} Note that the offsets represent the output of the deformation network. Multi-staged canonical spaces produce better image quality.}
\vspace{-5px}
\label{tab:ablation_hierarchy}
\end{table}

\begin{figure}
\centering
     \begin{subfigure}[t]{0.2\columnwidth}
         \centering
         \includegraphics[width=\columnwidth]{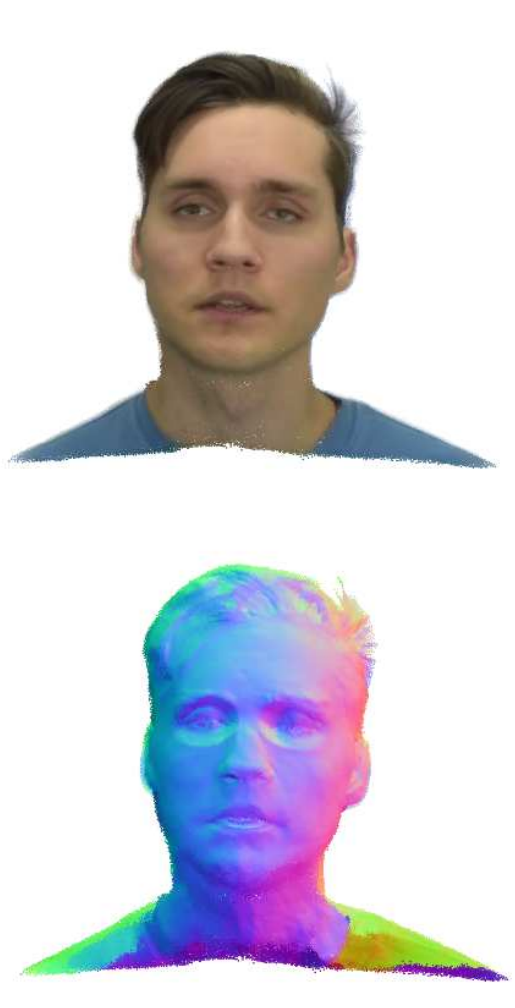}
         \caption{No Edit}
         \label{fig:no_edit}
     \end{subfigure}
     \begin{subfigure}[t]{0.22\columnwidth}
         \centering
         \includegraphics[width=\columnwidth]{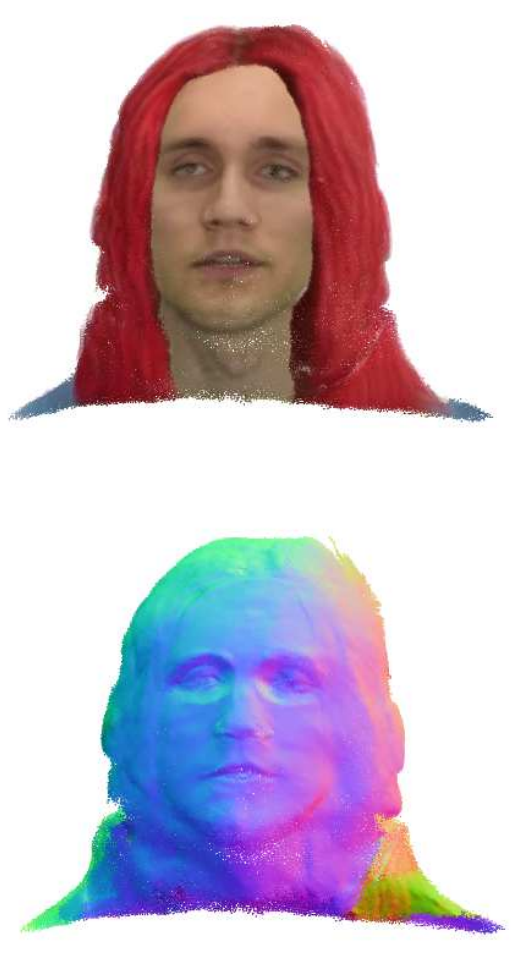}
         \caption{Hair, Nose}
         \label{fig:hair_nose}
     \end{subfigure}
     \begin{subfigure}[t]{0.22\columnwidth}
         \centering
         \includegraphics[width=\columnwidth]{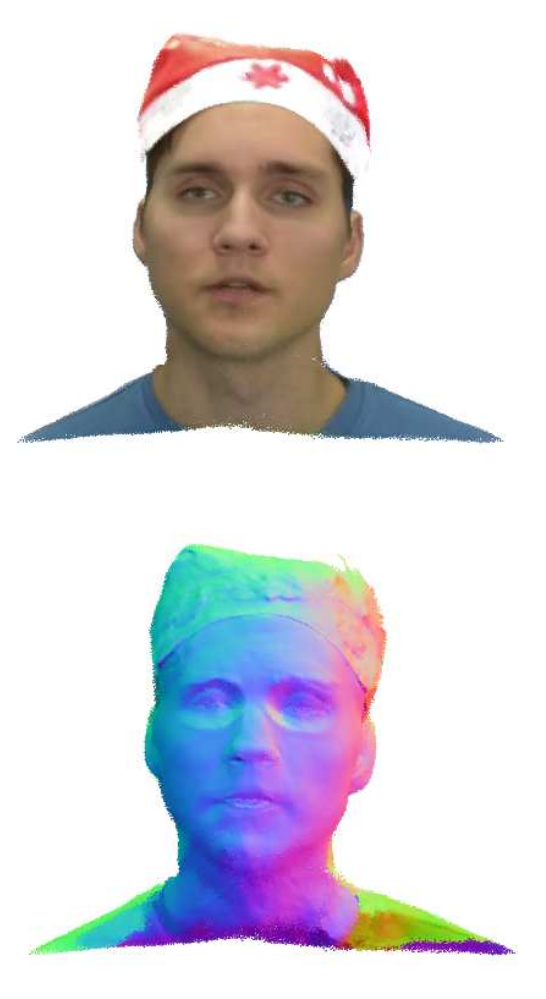}
         \caption{Hat, Mouth}
         \label{fig:hat_mouth}
     \end{subfigure}
     \begin{subfigure}[t]{0.22\columnwidth}
         \centering
         \includegraphics[width=\columnwidth]{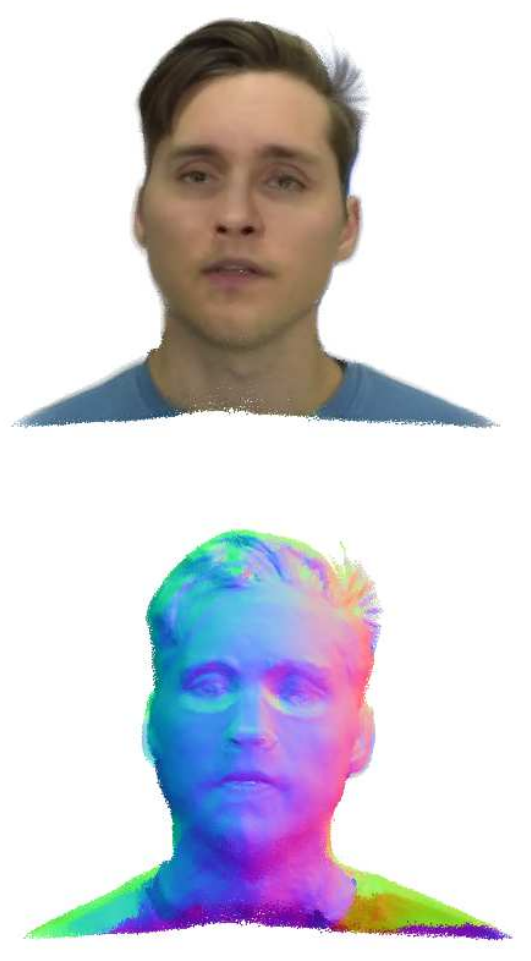}
         \caption{Eyes, Nose}
         \label{fig:eyes_noses}
     \end{subfigure}
     \vspace{-5px}
     \caption{\textbf{Multiple Composition.} PEGASUS generates the avatar with multiple face attributes.}
     \vspace{-5px}
\label{fig:multi_composition}
\end{figure}

\begin{figure}[t]
\includegraphics[trim={0 0 0 0},clip,width=\columnwidth]{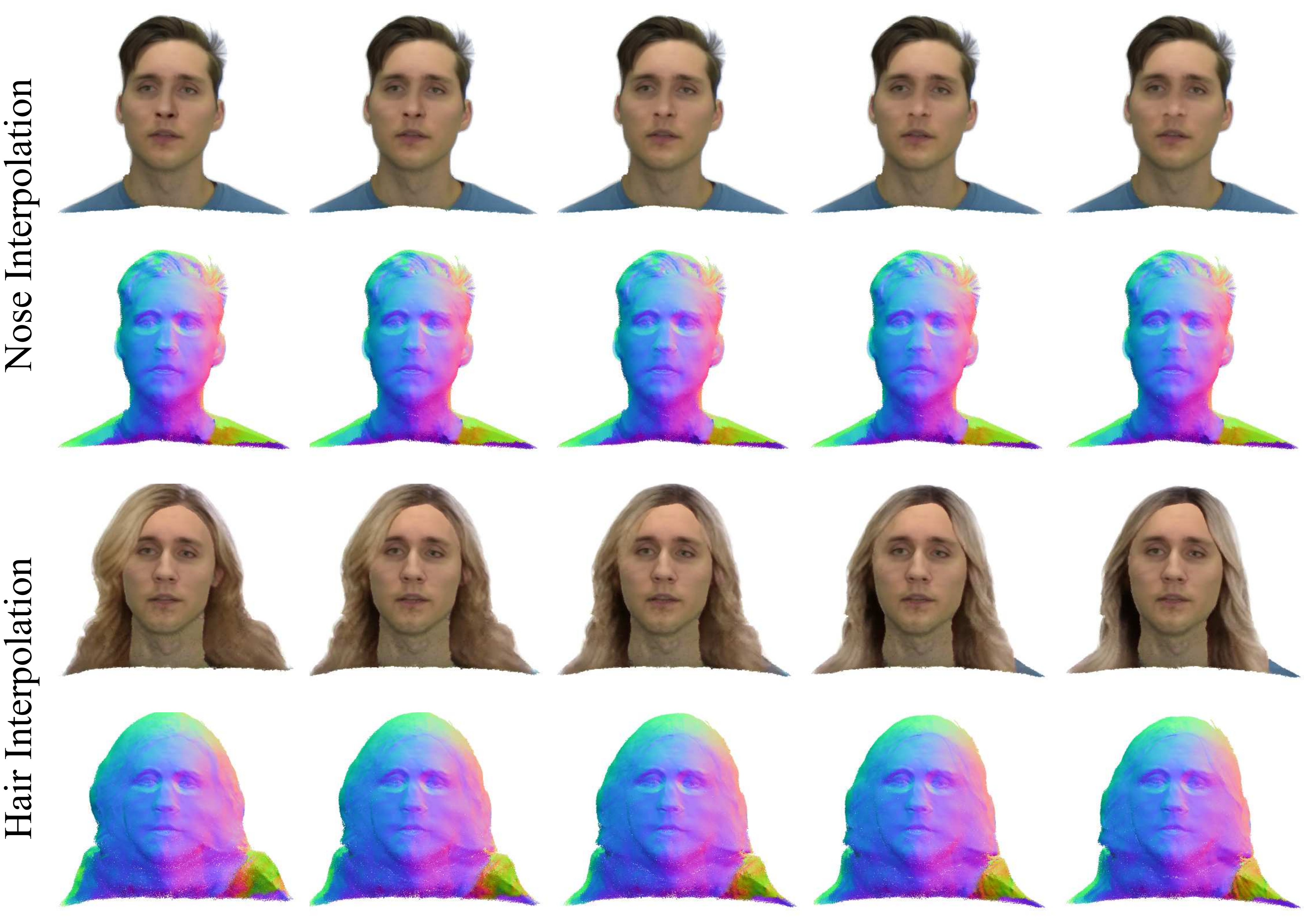}
\vspace{-15px}
\caption{\textbf{Person-specific Interpolation.} We interpolate the attribute latent code $\mathbf{z}^p, p \in \text{\{hair, nose\}}$ between two avatars.} 
\label{fig:person_specific_interpolation}
\vspace{-5px}
\end{figure}

\vspace{-3px}
\noindent \textbf{Multi-Stage Canonical Space.} 
We compare our multi-stage canonical space framework with the alternative framework with one- or two-stage by PointAvatar frameworks. 
For quantitative comparison on multi-stage canonical space, we use PSNR, SSIM, and LPIPS~\cite{zhang2018unreasonable} metrics. We evaluate them on unseen test sequences with novel head poses and facial expressions from all synthesized videos as shown in \figref{fig:synthesized_videos}. 
In \tabref{tab:ablation_hierarchy}, our multi-stage canonical space and point deformation outperforms the others stage deformation of all metrics.
\noindent \textbf{More Qualitative Results.} 
We further demonstrate the performance of our methods by showing the ability to control multiple parts as shown in \figref{fig:multi_composition}, and also by showing more interpolation results as shown in \figref{fig:person_specific_interpolation} and our Supp. Video.
\vspace{-3px}
\vspace{-12px}

\section{Discussion}
\vspace{-3px}
\label{sec:discussion}
We present a method for constructing personalized generative 3D face avatars from monocular video sources. Our compositional generative model enables disentangled controls to selectively alter the facial attributes of the target individual while preserving the identity. Notably, our personalized generative model is built exclusively from monocular videos, without relying on complex multi-view system setups. To achieve this goal, we present a method to construct a person-specific generative 3D avatar by building a synthetic video collection of the target identity with varying facial attributes, where the videos are synthesized by borrowing parts from diverse individuals from other monocular videos. We also show a zero-shot approach to achieve the same generative modeling more efficiently. For future research, building a more generative model to include multiple identities in a single model can be another exciting extension of our model. 

\noindent \textbf{Limitation.} As a limitation, the quality of our personalized avatar still does not reach the photo-realistic quality, showing noticeable artifacts. Also, due to the reliance on non-physical-based methods for generating the synthetic DB, our approach exhibits limitations in achieving physical accuracy. We show the failure cases and limitations of the synthetic DB generation in Supp. Mat.

\noindent \textbf{Acknowledgements.} 
We thank Inhee Lee for the fruitful discussion and Hyunwoo Cha for an essential role in collecting and processing in-the-wild videos. This work was supported by SNU Creative-Pioneering Researchers Program, NRF grant funded by the Korean government (MSIT) (No. 2022R1A2C2092724), and IITP grant funded by the Korean government (MSIT) (No.2022-0-00156, No.2021-0-01343). H. Joo is the corresponding author. %

{
    \small
    \setlength{\bibsep}{0pt}
    \bibliographystyle{abbrvnat}
    \bibliography{shortstrings, 11_references}
}

\ifarxiv \clearpage \appendix
\label{sec:appendix_cr}

\section{Synthetic DB Generation}
In this section, we provide further details of our synthetic database (DB) generation via part swapping, introduced in \secref{sec:synthetic} of our main manuscript.

\noindent \textbf{Hair.} We empirically find that removing the hair of the target subject is necessary before swapping the hair from the attribute DB. To create a bald head representation of the target individual, we utilize the Stable Diffusion~\cite{rombach2022high}, employing auto-generated mask images for this purpose. To generate the hair mask, we utilize an off-the-shelf face parsing network~\cite{yu2018bisenet, zllrunning_faceparsing}. We dilate the mask image using a kernel of size 20 from OpenCV~\cite{opencv_library}. Then, to generate an image of the target person with a bald head, we employ Stable Diffusion in conjunction with ControlNet~\cite{zhang2023adding}. The prompt to generate the bald head is ``bald, clean skin, smooth bald, small head, albedo." The negative prompt is ``hair, wrinkles, shadow, light reflection, tattoo, sideburns, facial hair, cartoonish, abstract interpretations, hat, head coverings." The examples are shown in \figref{fig:stable_diffusion_inpatining}. 

\noindent \textbf{Other Attributes.} Our goal is to synthesize the shape and appearance of the facial attribute from the attribute DB into the target individual as seamlessly as possible. To achieve this, we first render the avatar from an attribute DB into the same view, shape, and facial expressions as the target frame of the target individual's video, as described in Sec.~4.2 in our main manuscript. Subsequently, we acquire the mask of the rendered facial attribute by employing a face parsing network~\cite{yu2018bisenet, zllrunning_faceparsing} and then slightly enlarge it by applying the dilate function in OpenCV. We also perform the segmentation for the target individual's image to acquire the mask of the target facial attribute by utilizing the face parsing network~\cite{yu2018bisenet, zllrunning_faceparsing}, where the target facial part is subsequently ``removed'' via inpainting by employing the Fast Marching Method~\cite{telea2004image}. This process can be considered as a similar process of ``bald head synthesis'' before integrating the desired facial part from the attribute source. Finally, we seamlessly integrate the facial attribute from the attribute avatar into the target individual using Poisson blending~\cite{perez2023poisson}. Examples of nose and mouth synthesis employing this technique are illustrated in \figref{fig:poisson_blending_inpainting}.

\begin{figure}[t]
\includegraphics[trim={0 0 0 0},clip,width=1.0\columnwidth]{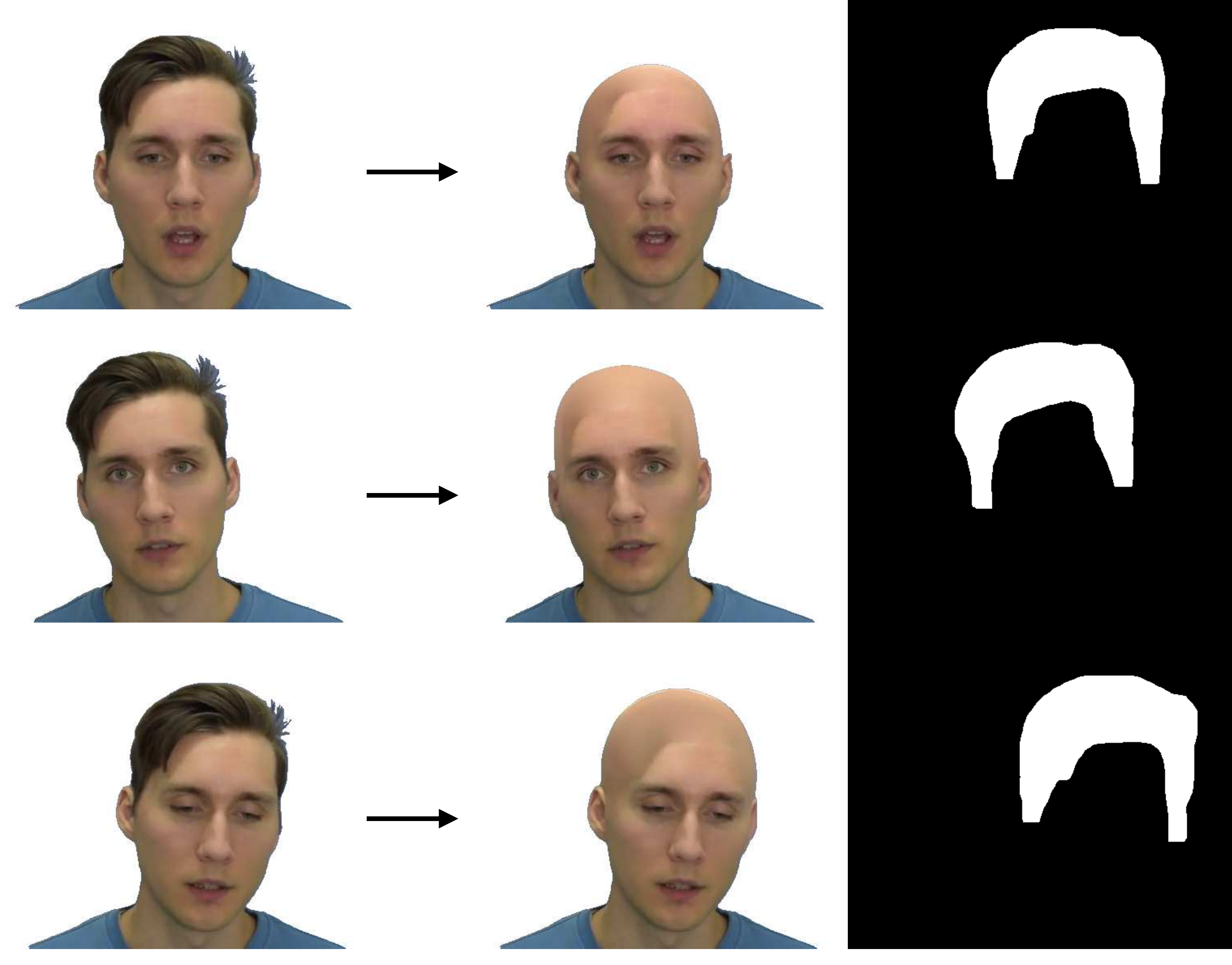}
\caption{\textbf{Stable Diffusion Inpainting.} We leverage Stable Diffusion~\cite{rombach2022high} and ControlNet~\cite{zhang2023adding} to remove the target's hair and make it bald, in order to synthesize different hair. The automatically generated mask images represent the area designated for inpainting.}
\label{fig:stable_diffusion_inpatining}
\end{figure}

\begin{figure}[t]
\includegraphics[trim={0 0 0 0},clip,width=1.0\columnwidth]{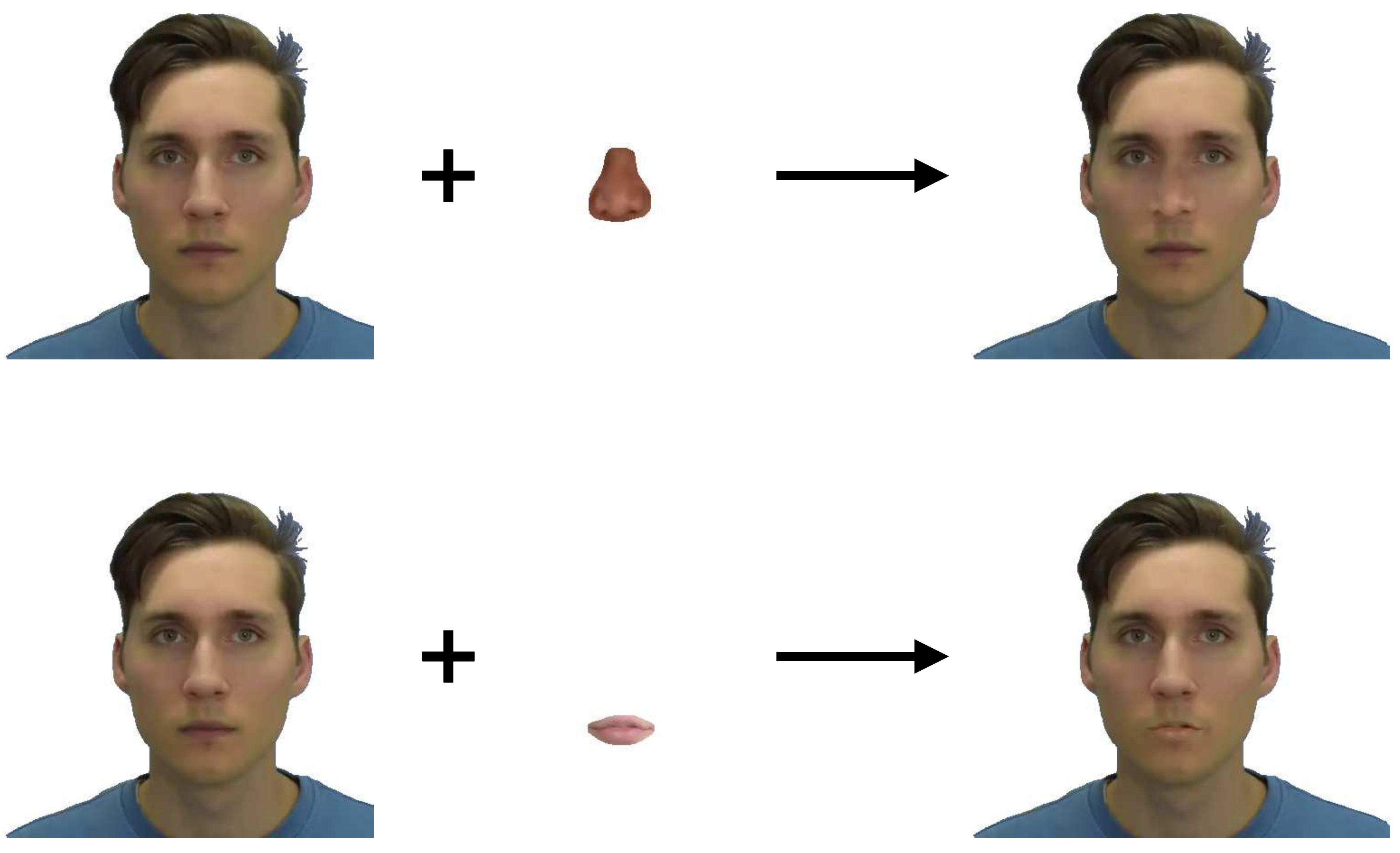}
\caption{\textbf{Poisson Blending Inpainting.} We use Poisson blending~\cite{perez2023poisson} to synthesize the facial attribute and the target's face.}
\label{fig:poisson_blending_inpainting}
\end{figure}

\noindent \textbf{Tracking and Masking.}
To extract FLAME parameters from images, along with their corresponding camera parameters, we utilize the DECA model ~\cite{feng2021learning}. When FLAME parameters are directly extracted using the DECA model, we notice that the head pose estimation is noisy and jittery, particularly in the frames where the eyes in the original images are blinking. To improve the FLAME parameter estimation quality, following the similar process of PointAvatar~\cite{zheng2023pointavatar}, we apply an optimization procedure to align the 2D projection of FLAME's facial landmarks with the detection outputs of an off-the-shelf 2D facial landmark detector~\cite{bulat2017far}. This optimization process is based on the assumption that the quality of the 2D landmark detection is more precise. We minimize the point-wise distance between the landmark obtained from FLAME and the 2D facial landmark to optimize the shape, pose, and camera parameters. Different from PointAvatar’s approach, instead of using a singular translation vector for each video, we employ a unique vector for every image frame in scenarios involving in-the-wild video tracking.

To create the foreground mask image, we leverage an off-the-shelf background matting network~\cite{ke2022modnet} to obtain the portrait mask images from the videos. We use the face parsing network~\cite{yu2018bisenet, zllrunning_faceparsing} to obtain part segmentations of the faces and leverage SegmentAnything model~\cite{kirillov2023segany, liu2023grounding} for segmenting head accessories.

\section{Postprocessing of Zero-Shot Transfer}\label{sec:post_processing_zero_shot}
We provide further details of the \eqnref{eq:zero_shot_naive_pcd} in our main manuscript, which is the process of combining the subsets of point clouds from both avatars. In short, the zero-shot process is performed via three steps: (1) naive composition after segmentation by introducing additional point clouds for the missing region; (2) optimization by aligning facial landmarks for better alignment; and (3) color blending for the added points for seamless outputs. 

\noindent \textbf{Obtaining the Additional Part from the Source Human.} 
We use the estimated segmentation masks of the face attribute $\chi_{\phi}$ and $\chi_{th}$ that can be controlled via latent code $\mathbf{z}$ to select the \emph{target human}'s point cloud except for the facial attribute $\chi_{th} = 0$ and \emph{source human}'s point cloud that includes the facial attribute $\chi_{\phi} = 1$:
\begin{align}\label{eq:zero_shot_naive_pcd_recap}
    P_{\text{naive}} = \{ \mathbf{x}_{\phi}^{d, i} \}_{\chi^{i}_{\phi} = 1} \cup \{ \mathbf{x}_{th}^{d, i} \}_{\chi^{i}_{th} = 0}
\end{align}
When we remove the facial attribute from the \emph{target human} and bring in the facial attribute from the \emph{source human}, it creates an empty space between the two point clouds. To fill this missing region, as shown in \figref{fig:zero_shot_optim1}, we bring in additional parts from the \emph{source human}. Formally, this can be represented as follows:
\begin{align}\label{eq:zero_shot_naive_additional_pcd}
    P_{\text{naive w/ add}} = P_{\text{naive}} \cup \{ \mathbf{x}_{\phi}^{d, i} \}_{\chi^{i}_{\phi, \text{add}} = 1}
\end{align}
To create the additional segmentation mask $\chi_{\phi,~\text{add}}^i$, we borrow the knowledge from the FLAME~\cite{li2017flame} by leveraging $k$-nearest neighbor $\mathcal{N}$. $\mathcal{N}_k(P_1, P_2)$ denotes the $k$-nearest neighbors in $P_2$ for each point in $P_1$. $\arg\min\mathcal{N}_k(P_1, P_2)$ represents the indices of the $k$-nearest neighbors from points in $P_1$ to points in $P_2$~\cite{ravi2020pytorch3d}. We omit the subscript $k$ when $k=1$. 

\begin{figure}[t]
\includegraphics[trim={0 0 0 0},clip,width=\columnwidth]{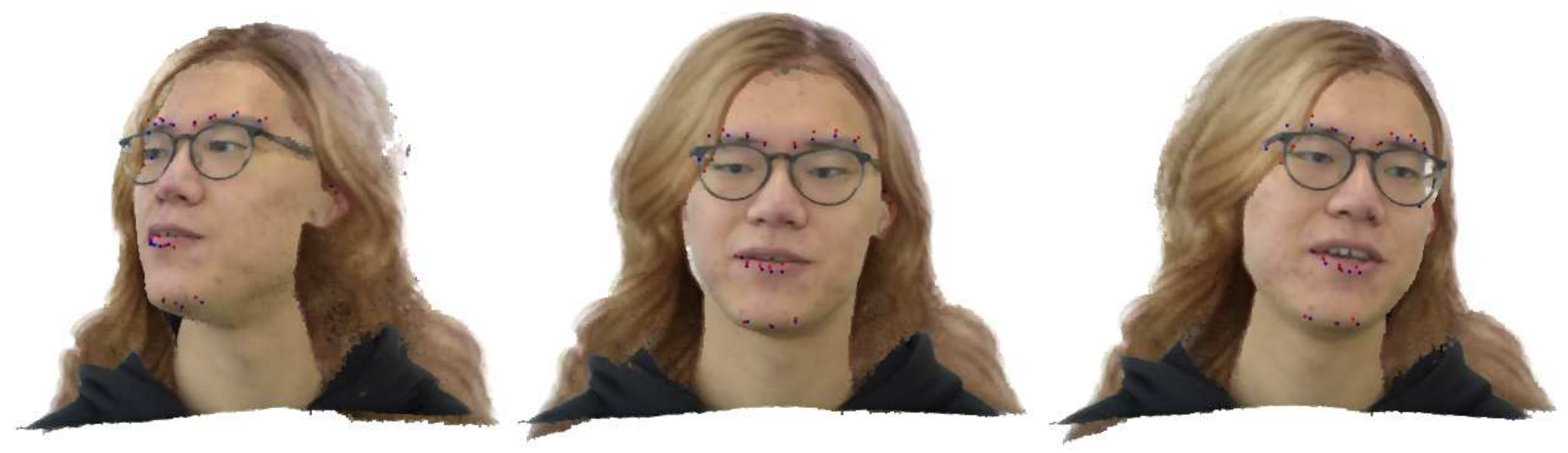}
\caption{\textbf{Zero-Shot Landmarks for Optimization.} The red dot represents our personalized generative model's $k$-nearest neighbor of 3D Landmarks from FLAME keypoints, and the blue dot represents the target's $k$-nearest neighbor of 3D Landmarks from FLAME keypoints.}
\label{fig:zero_shot_optimization}
\end{figure}
\begin{figure}[t]
\centering
     \begin{subfigure}[t]{0.30\columnwidth}
         \centering
         \includegraphics[width=\columnwidth]{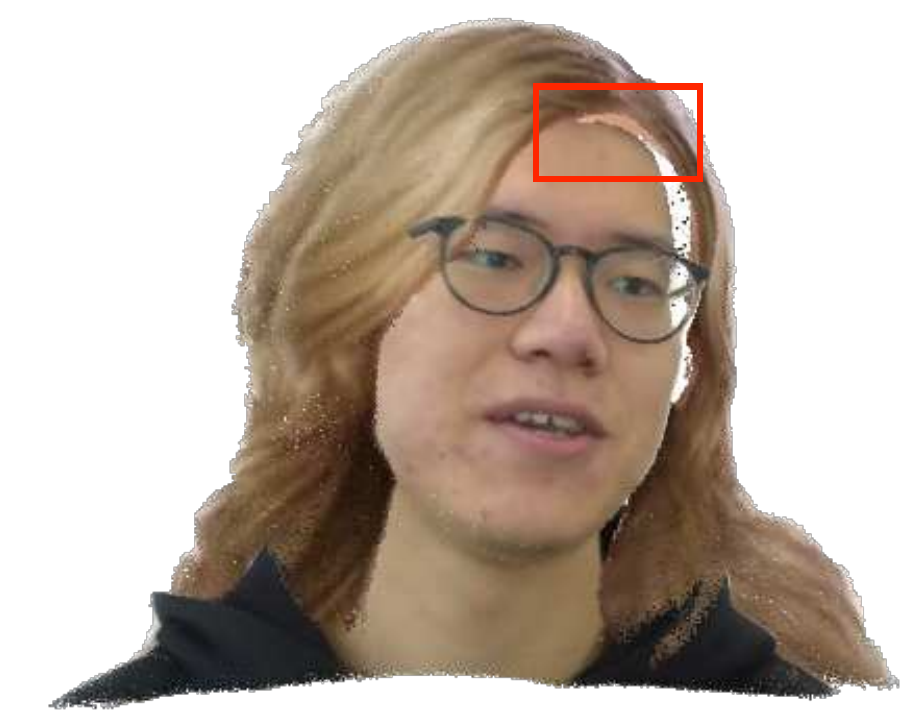}
         \caption{Before Optimize.}
         \label{fig:zero_shot_optim1}
     \end{subfigure}
     \begin{subfigure}[t]{0.28\columnwidth}
         \centering
         \includegraphics[width=\columnwidth]{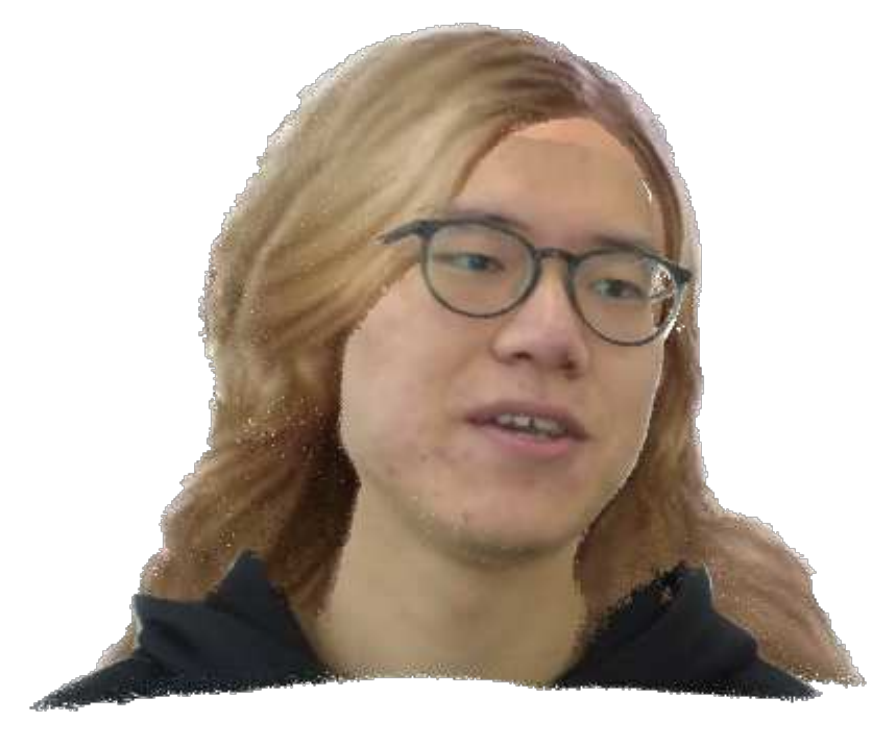}
         \caption{After Optimize.}
         \label{fig:zero_shot_optim2}
     \end{subfigure}
     \begin{subfigure}[t]{0.28\columnwidth}
         \centering
         \includegraphics[width=\columnwidth]{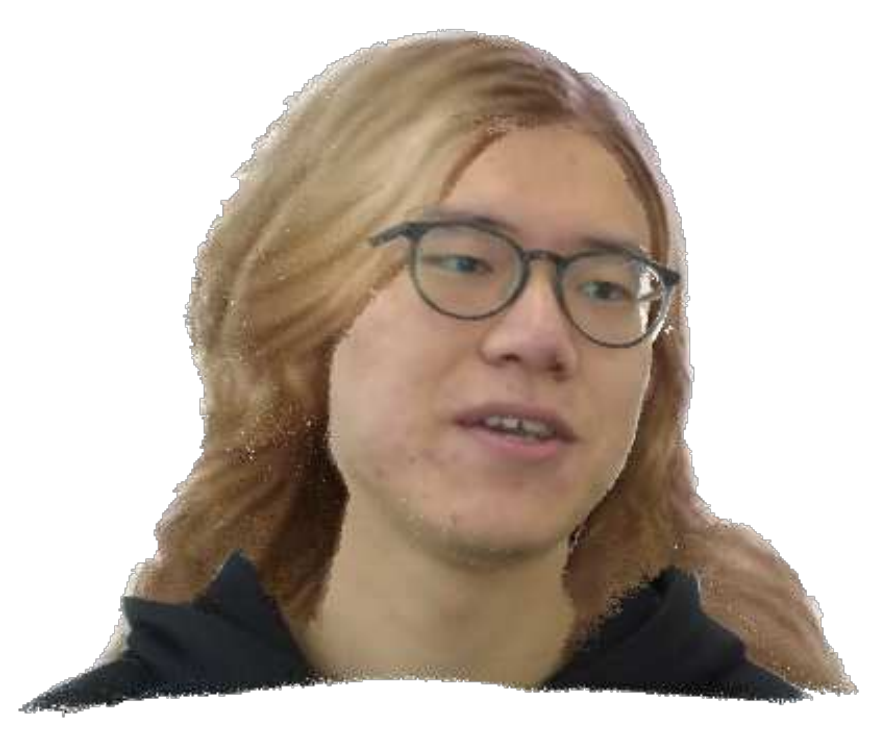}
         \caption{Color Blending.}
         \label{fig:zero_shot_optim3}
     \end{subfigure}
     \hfill
     \caption{\textbf{Zero-Shot Optimization Steps.} The red box represents the additional part to fill the empty space. Through zero-shot modeling, we generate an avatar with a high-quality and reasonable appearance in three stages of post-processing in \secref{sec:post_processing_zero_shot}. }
\label{fig:zero_shot_optim}
\end{figure}

Note that $\{ \mathbf{x}_\phi^{d,i} \}_{\chi^i_{\phi, \text{add}}=1}$ denotes the additional point clouds from the \emph{source human} to fill the gaps between the \emph{source human} and \emph{target human} because of the exception of \emph{target human}'s attribute, as shown in the red box of \figref{fig:zero_shot_optim1}. To create $\chi_{\phi, \text{add}}^{i}$, We exclude the vertices from the FLAME vertices $\mathbf{x}_{th}^{\text{FLAME}}$ that are not associated with the additional part by using $\chi_{th}$ and the back of the head part of the FLAME that we designate. We denote the mask cue for obtaining FLAME corresponding to the additional part as $\chi_{th, \text{add}}^\text{FLAME}$. 
\begin{equation}\label{eq:zero_shot_selected_flame_gap}
     \mathbf{x}_{th,\text{add}}^{\text{FLAME}} = \{\mathbf{x}_{th}^{\text{FLAME}}\}_{\chi_{th, \text{add}}^\text{FLAME}}
\end{equation}
We apply $\mathcal{N}$ to $\mathbf{x}_{\phi}^{d}$ and $\mathbf{x}_{th,\text{add}}^{\text{FLAME}}$ to obtain the nearest neighbor of \emph{source human}. To create the additional part only, we use $(1-\chi_{\phi})$ except for \emph{source human}'s attribute.
\begin{equation}\label{eq:additional_sh}
    \chi_{\phi,\text{add}} = (1-\chi_{\phi}) \circ {\arg\min}\mathcal{N}_k(\mathbf{x}_{th,\text{add}}^{\text{FLAME}}, \mathbf{x}_{\phi}^{d}),
\end{equation}
where $\circ$ represents the Hadamard product. We use $k=2000$ to generate the additional point clouds as described in \figref{fig:zero_shot_optim1}. 

\noindent \textbf{Optimization Step.} After the naive composition, there is still a gap between the \emph{source human}'s face attribute and \emph{target human}'s other parts because of the misalignment of the subject-specific FLAME canonical space, as shown in \figref{fig:zero_shot_optim1}. 
To solve this issue, we apply the optimization process to minimize the distance between the \emph{source human} and \emph{target human}. To obtain the landmark points, we apply the $k$-nearest neighbor function between the landmarks of deformed FLAME vertices~\cite{feng2021learning} and $\mathbf{x}^d$ as follows:
\begin{equation}
    \mathbf{x}^{\text{landmarks}} = \mathcal{N}(\mathbf{x}^{\text{FLAME landmarks}}, \mathbf{x}^d)
\end{equation}
We leverage the distance between the \emph{source human}'s 3D landmark points and the \emph{target human}'s 3D landmark points as shown in \figref{fig:zero_shot_optimization}. 
\begin{equation}
    \mathbf{d}_1 = \| \mathbf{x}_{th}^{\text{landmarks}} - \mathbf{x}_{\phi}^{\text{landmarks}} \|_2^2
\end{equation}
Furthermore, we calculate the squared distances between points in the additional \emph{source human} part, denoted as $\{\mathbf{x}_{\phi}^d\}_{\chi_{\phi,\text{add}}=1}$, and points in the \emph{target human}, represented by $\{\mathbf{x}_{th}^d\}_{\chi_{th}=0}$, from the $k$-nearest neighbors. For simplicity, the superscript $i$ is omitted.
\begin{equation}
    \mathbf{d}_2 = \| \mathcal{N}(\{\mathbf{x}_{\phi}^d\}_{\chi_{\phi,\text{add}}=1}, \{\mathbf{x}_{th}^d\}_{\chi_{th}=0}) \|_2^2
\end{equation}
We optimize the learnable angle-axis rotation vector $R \in \mathbb{R}^3$ and translation vector $t \in \mathbb{R}^3$ to minimize the distance $\mathbf{d} = \mathbf{d}_1 + \mathbf{d}_2$ by Adam optimizer~\cite{kingma2014adam}. Note that we apply the rotation and translation vector at the subject-specific FLAME-canonical space.
\begin{equation}
    \mathbf{x}_{\phi, \text{moved}}^{fc} = R \cdot \{\mathbf{x}_{\phi}^{fc}\} + t
\end{equation}
We obtain the moved \emph{source human}'s point cloud $\mathbf{x}_{\phi, \text{moved}}^d$ from $\mathbf{x}_{\phi, \text{moved}}^{fc}$ by \eqnref{eq:fc_to_bs}. As a consequence, the optimized point cloud is represented as follows:
\begin{equation}
    P_{\text{optim}} = \{ \mathbf{x}_{\phi, \text{moved}}^{d} \}_{\chi_{\phi} \circ \chi_{\phi, \text{add}} = 1} \cup \{ \mathbf{x}_{th}^{d} \}_{\chi_{th} = 0} 
\end{equation}
The optimized rendering result is shown in \figref{fig:zero_shot_optim2}.

\noindent \textbf{Blending Step.} To generate a natural rendering of the additional part, denoted as $\{ \mathbf{x}_\phi^{d} \}_{\chi_{\phi, \text{add}}=1}$, we leverage the feature information from the \emph{target human} using the $k$-nearest neighbor. 
\begin{equation}
    \mathbf{c}_{\text{add}}^d = {\arg\min}\mathcal{N}(\mathbf{x}_{th}, \{\mathbf{x}_{\phi}^d\}_{\chi_{\phi,\text{add}}=1}) \circ \mathbf{c}_{th}^d
\end{equation}
\begin{equation}
    \mathbf{n}_{\text{add}}^d = {\arg\min}\mathcal{N}(\mathbf{x}_{th}, \{\mathbf{x}_{\phi}^d\}_{\chi_{\phi,\text{add}}=1}) \circ \mathbf{n}_{th}^d
\end{equation}
The RGB and normal of the additional part come from the target human, so we obtain the naturally blended avatar through the zero-shot model. The natural blended results are shown in \figref{fig:zero_shot_optim3}.

\begin{figure}[t]
\includegraphics[trim={0 0 0 0},clip,width=1.0\columnwidth]{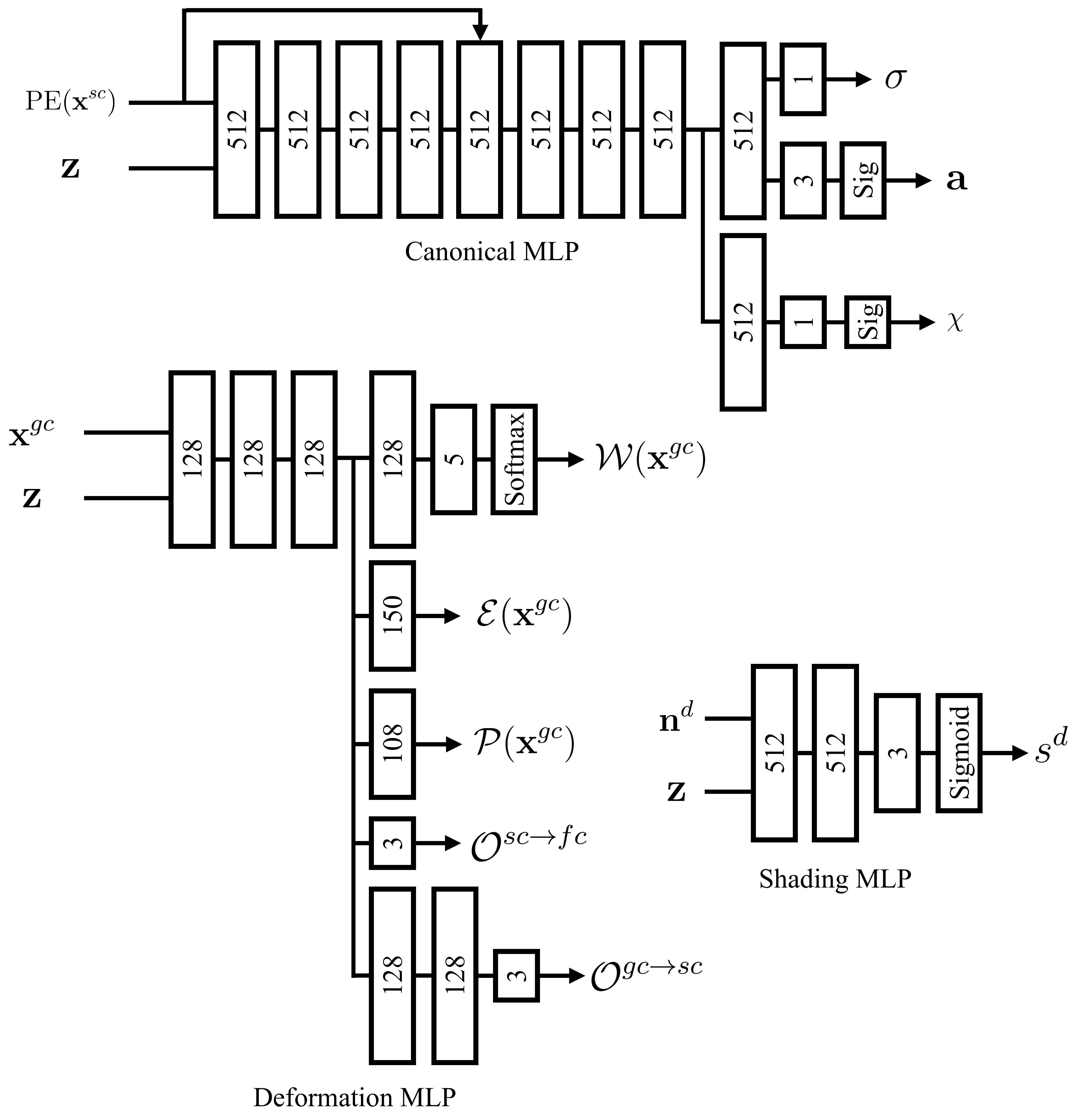}
\caption{\textbf{Network Architecture of PEGASUS.} In PEGASUS, the latent code $\mathbf{z}$ serves as a condition for all the MLPs.}
\label{fig:network_architecture}
\end{figure}

\section{Implementation Details}
\noindent \textbf{Network Architecture}
In \figref{fig:network_architecture}, we show the network architecture of PEGASUS. Following PointAvatar~\cite{zheng2023pointavatar}, we leverage ReLU activation function~\cite{nair2010rectified} for shading MLP, and Softplus activation function for canonical and deformation MLP for every layer. \emph{Sig} denotes the sigmoid function in \figref{fig:network_architecture}. Different from PointAvatar, we use an additional layer to output segmentation cues $\chi$ in canonical MLP. Also, we use two layers of MLP to create subject-specific canonical offset $\mathcal{O}^{gc \rightarrow sc}$

\noindent \textbf{Loss Functions.} The total loss for PEGASUS is defined as follows:
\begin{equation}\label{eq:PointAvatar_loss}
\begin{split}
    \mathcal{L} & = \lambda_{\text{rgb}}\mathcal{L}_{\text{rgb}} + \lambda_{\text{mask}}\mathcal{L}_{\text{mask}} + \lambda_{\text{FLAME}}\mathcal{L}_{\text{FLAME}} + \lambda_{\text{vgg}}\mathcal{L}_{\text{vgg}}\\
    & + \lambda_{\text{normal}}\mathcal{L}_{\text{normal}} + \lambda_{\text{seg}}\mathcal{L}_{\text{seg}} + \lambda_{\mathbf{z}~\text{reg}}\mathcal{L}_{\mathbf{z}~\text{reg}}
\end{split}
\end{equation}
We leverage the loss functions from the facial implicit representations from monocular inputs~\cite{zheng2023pointavatar, zheng2022imavatar} as follows: 
\begin{gather}
    \mathcal{L}_\text{rgb} = \| c - c^\text{GT} \| \\
    \mathcal{L}_\text{mask} = \| M - M^\text{GT} \| \\
    \mathcal{L}_\text{vgg} = \| F_\text{vgg}(c) - F_\text{vgg}(c^\text{GT}) \| \\
    \begin{split}
        \mathcal{L}_\text{FLAME} = {\frac{1}{N}}&\sum_{i=1}^{N}( \lambda_{e}\|{\mathcal{E}}_{i}-{\hat{\mathcal{E}}}_{i}\|_{2} \\
        & +\lambda_{p}\|\mathcal{P}_{i}-{\hat{\mathcal{P}}}_{i}\|_{2} \\
        & +\lambda_{w}\|\mathcal{W}_{i}-{\hat{\mathcal{W}}}_{i}\|_{2}) 
    \end{split}
\end{gather}
Following PointAvatar, $c$ and $c^{GT}$ denote the color of the rendering images from PEGASUS and ground-truth color. $M$ denotes the mask from PEGASUS obtained by $\mathbf{m}_\text{pix} = \sum_i \alpha_i \mathbf{T}_i$. $F_\text{vgg}(\cdot)$ represent the features of pretrained VGG network~\cite{johnson2016perceptual, simonyan2014very}. $\mathcal{E}, \mathcal{P}, ~\text{and} ~\mathcal{W}$ are the pseudo ground truth of the $k$-nearest neighbor vertices of the FLAME~\cite{li2017flame}. Note that our method, PEGASUS, does not predict the shape blendshapes basis $\mathcal{S}$, directly using the $k$-nearest neighbor vertices of the FLAME.

Given ground-truth object mask $M_\text{seg}^{GT}$ and the predicted segmentation cues $\chi^{d}$, the rendered color of the segmented point cloud represents $\mathcal{R}(\chi^{d} \circ \mathbf{x}^d)$. The segmentation loss is defined as:
\begin{equation}\label{eq:segmentation_loss}
    \mathcal{L}_\text{seg} = \text{BCE}(\mathcal{R}(\chi^{d} \circ \mathbf{x}^d), M_\text{seg}^{GT})
\end{equation}
$\text{BCE}$ represent the Binary Cross-Entropy loss. $\mathcal{R}$ is the alpha composition rendering function. We leverage the alpha composition function of PyTorch3D~\cite{ravi2020pytorch3d} to render the predicted segmentation cues. 

We adopt the normal loss to encourage high-fidelity geometry and texture as follows:
\begin{equation}
    \mathcal{L}_\text{normal} = \| \mathbf{n} - \mathbf{n}^d\|
\end{equation}
We generate the pseudo ground truth normal $\mathbf{n}$ from the $V^{tp}$ and the avatar trained with a single identity of each $V^{db}$. 
We apply the regularization of latent code to be close to zero.
\begin{equation}
    \mathcal{L}_{\mathbf{z}~\text{reg}} = \| \mathbf{z} \|
\end{equation}

\noindent \textbf{Training Strategy}
We train PEGASUS in two stages. In the first stage, we only use the target individual from $V^{tp}$ for training. In this way, the initial point cloud is deformed from a sphere to have a reasonable face shape. In the second stage, we continue training using all part-swapped videos from $\hat{V}_i^{tp}$. We have empirically found that this two-stage training shows more reliable training. In all of our experiments, we start the second stage from the 10th epoch, using 1600 point clouds.

\section{More Ablation Study}
\noindent \textbf{Multi-stage Canonical Spaces.} In \figref{fig:ablation_offset} and \tabref{tab:ablation_hierarchy}, our multi-stage canonical space and point deformation method outperforms the best metrics and quality compared to other approaches. Notably, as an example of \figref{fig:ablation_offset}, the closest high-quality reconstruction to the Ground Truth (GT) is achieved by the three-stage approach. 

\begin{figure}
\centering
    \begin{subfigure}[t]{0.2\columnwidth}
         \centering
         \includegraphics[width=\columnwidth]{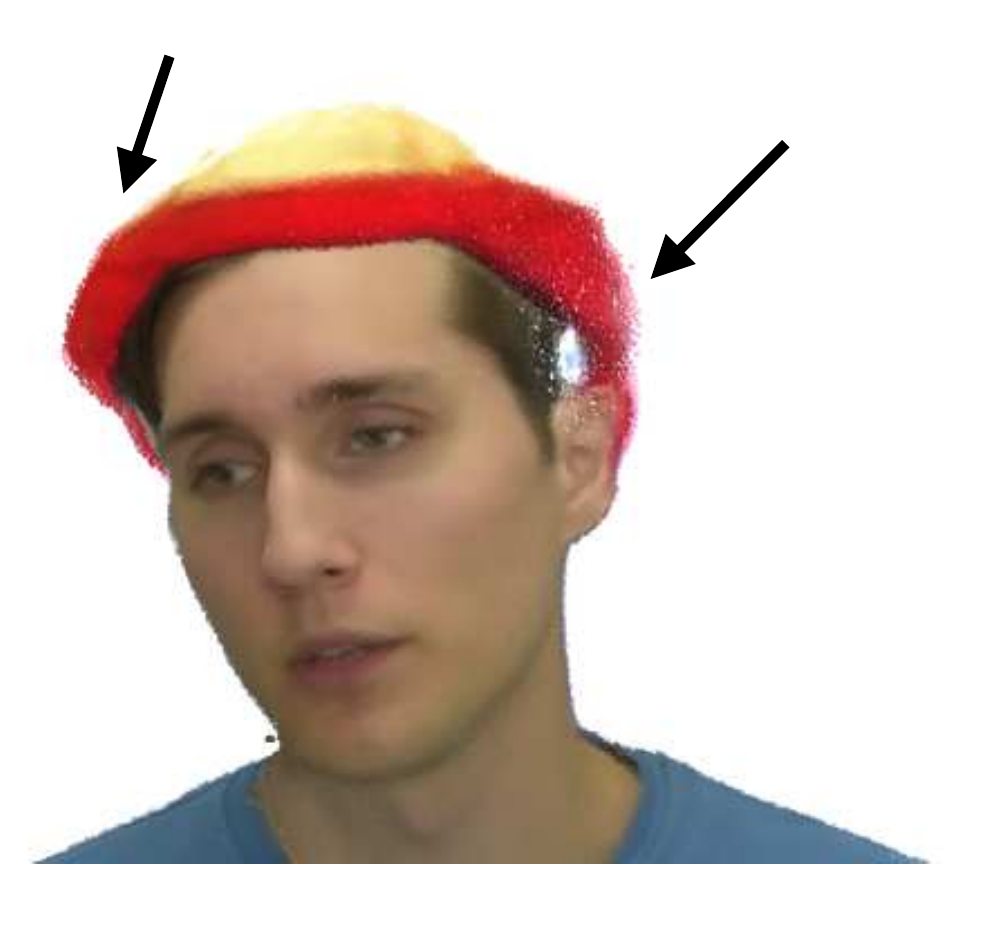}
         \caption{1-Stage}
         \label{fig:one_stage}
     \end{subfigure}
     \begin{subfigure}[t]{0.22\columnwidth}
         \centering
         \includegraphics[width=\columnwidth]{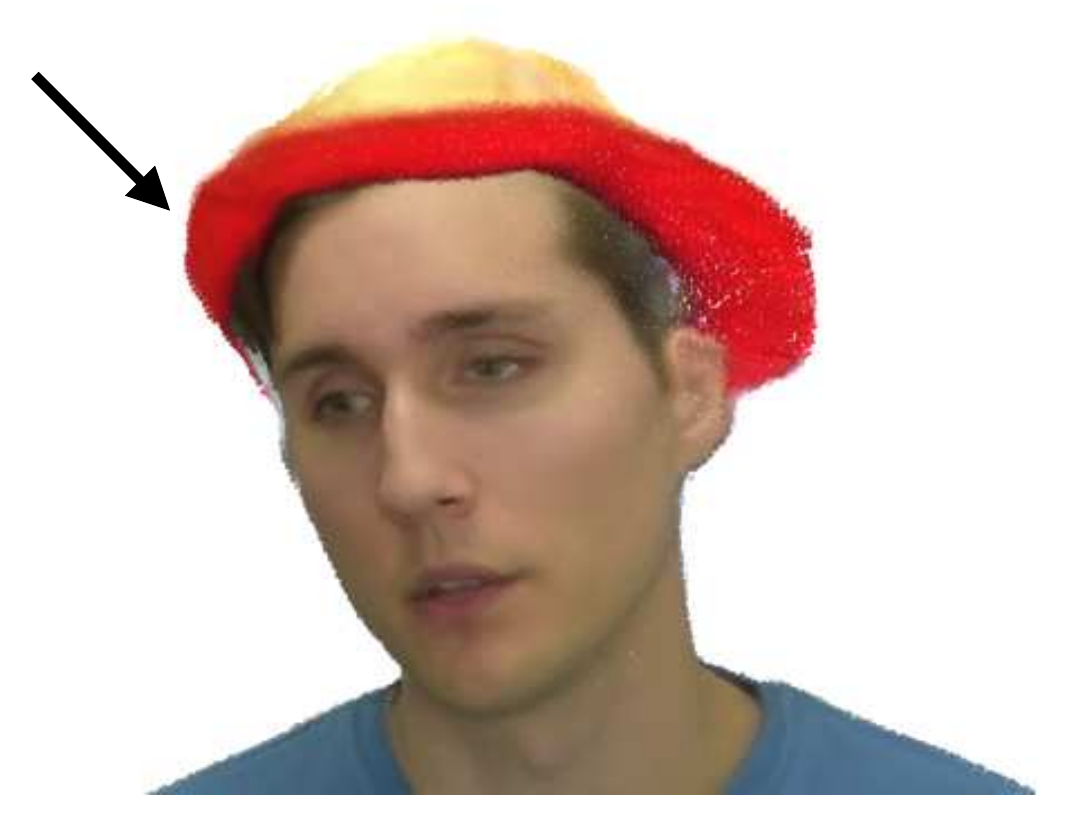}
         \caption{2-Stages}
         \label{fig:two_stages}
     \end{subfigure}
     \begin{subfigure}[t]{0.22\columnwidth}
         \centering
         \includegraphics[width=\columnwidth]{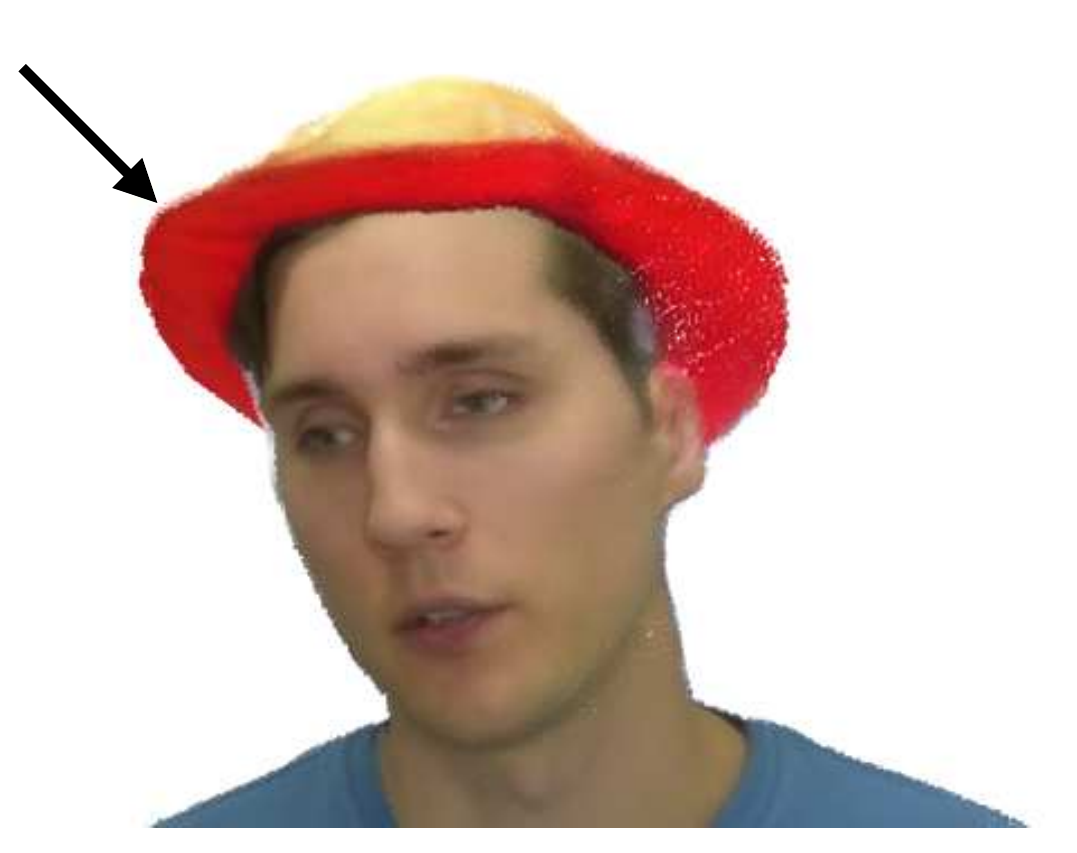}
         \caption{3-Stages}
         \label{fig:three_stages}
     \end{subfigure}
     \begin{subfigure}[t]{0.22\columnwidth}
         \centering
         \includegraphics[width=\columnwidth]{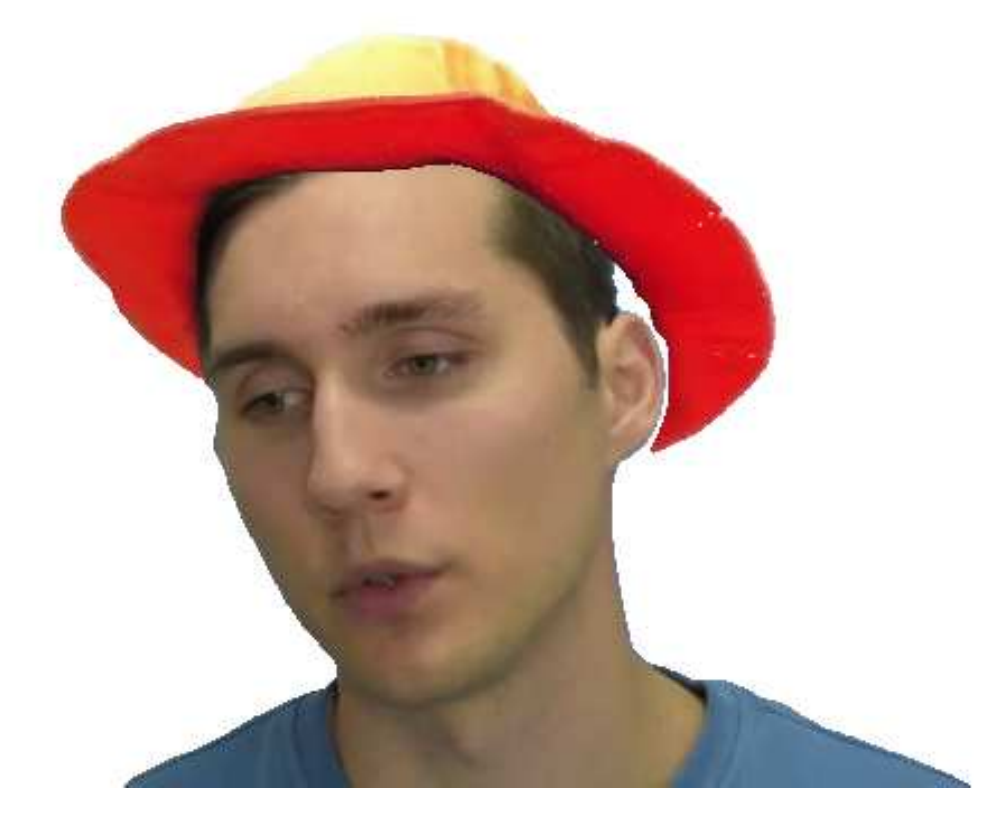}
         \caption{GT}
         \label{fig:gt}
     \end{subfigure}
     \hfill
     \caption{\textbf{Ablation: reconstruction by offsets $\mathcal{O}$.} Our three-stage canonical space framework creates reasonable and accurate reconstruction.}
\label{fig:ablation_offset}
\end{figure}
\begin{figure}
\centering
    \begin{subfigure}[t]{0.30\columnwidth}
         \centering
         \includegraphics[width=\columnwidth]{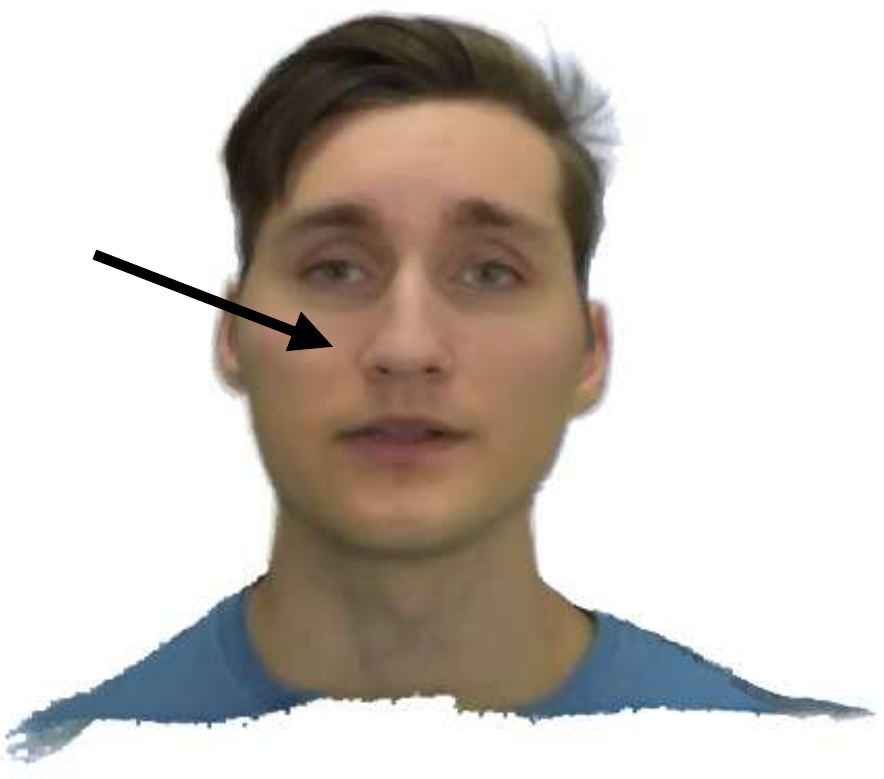}
         \caption{Default Nose}
         \label{fig:random_multi_stages1}
     \end{subfigure}
     \begin{subfigure}[t]{0.30\columnwidth}
         \centering
         \includegraphics[width=\columnwidth]{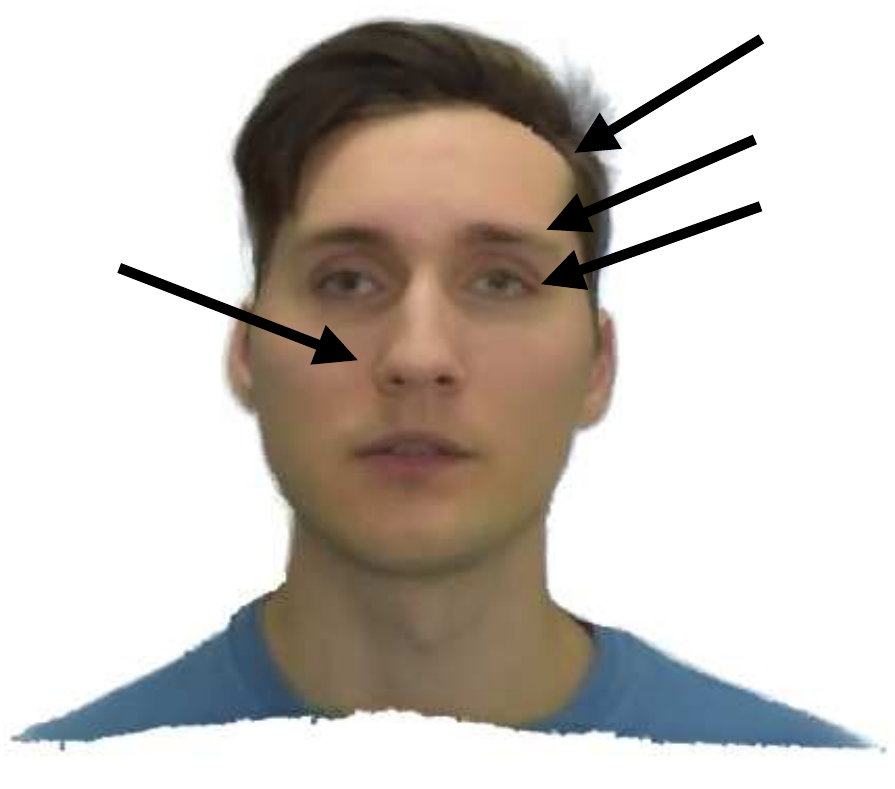}
         \caption{Two Stages}
         \label{fig:random_multi_stages2}
     \end{subfigure}
     \begin{subfigure}[t]{0.30\columnwidth}
         \centering
         \includegraphics[width=\columnwidth]{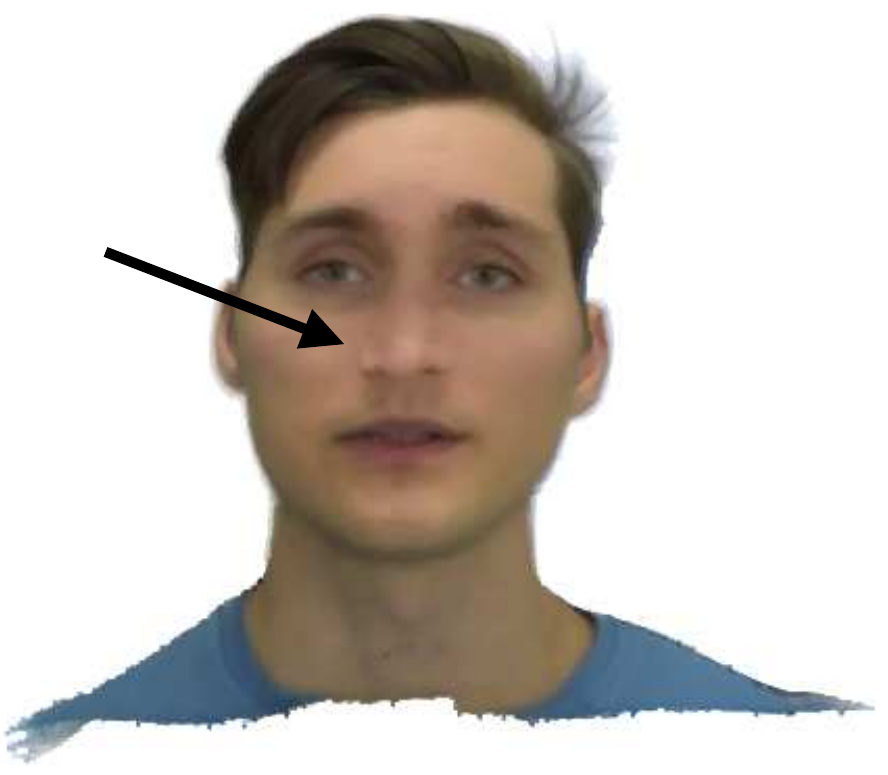}
         \caption{Ours}
         \label{fig:random_multi_stages3}
     \end{subfigure}
     \caption{\textbf{Ablation: random sampling by offsets $\mathcal{O}$.} Multi-stage canonical spaces enhance disentangled nose generation.}
\label{fig:multistage}
\end{figure}

Additionally, we randomly sample the nose latent codes to utilize two and three-stage baselines in \figref{fig:multistage}. \figref{fig:multistage} shows that the original architecture of PointAvatar fails to disentangle the target attribute. \figref{fig:multistage} demonstrates that the generic canonical space is necessary to generate disentangled facial attributes from random sampling while preserving individual identity. 

\begin{figure}
\centering
     \begin{subfigure}[b]{0.23\textwidth}
         \centering
         \includegraphics[width=\textwidth]{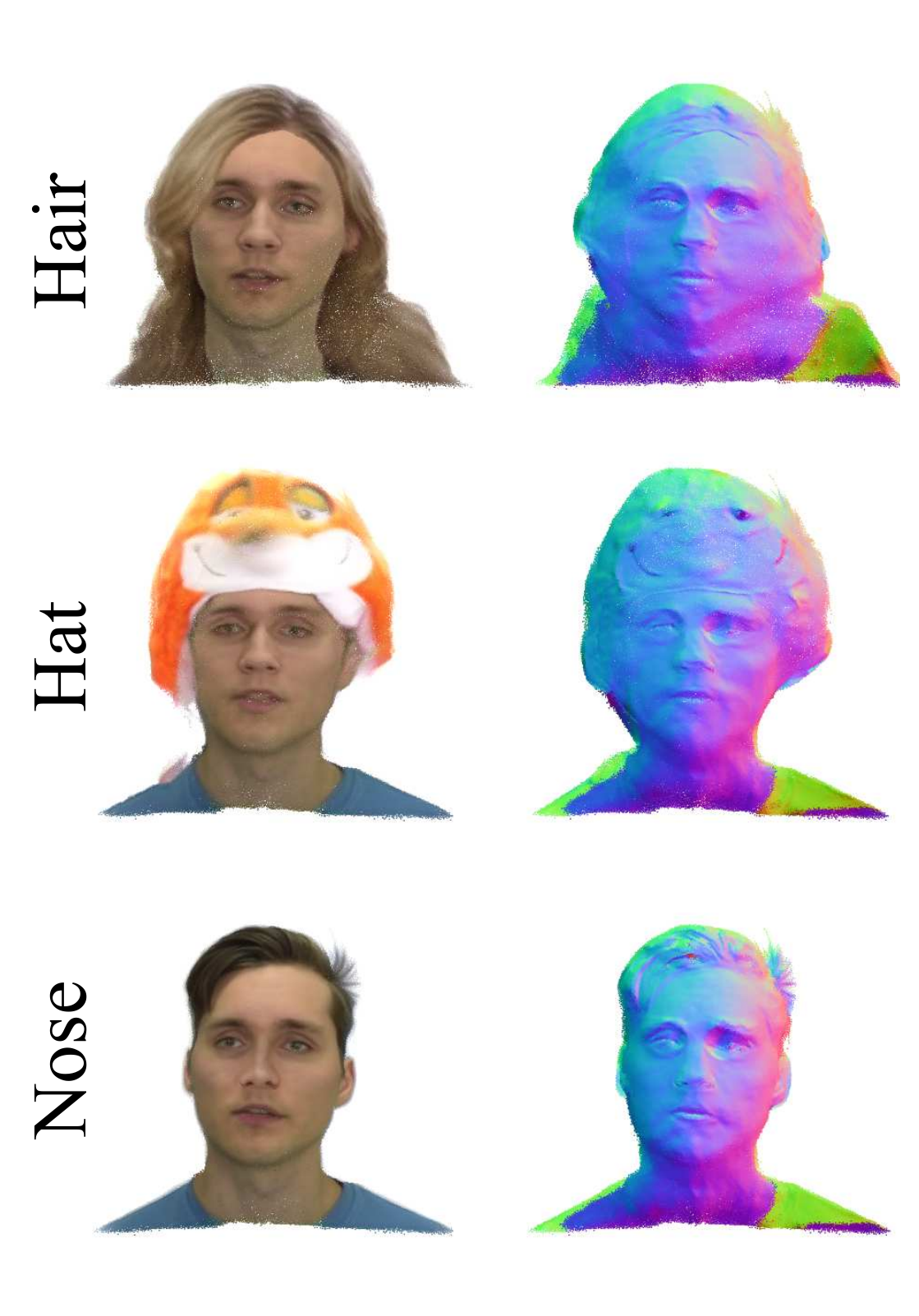}
         \caption{Without Normal Loss}
         \label{fig:face_before_resample}
     \end{subfigure}
     \begin{subfigure}[b]{0.2\textwidth}
         \centering
         \includegraphics[width=\textwidth]{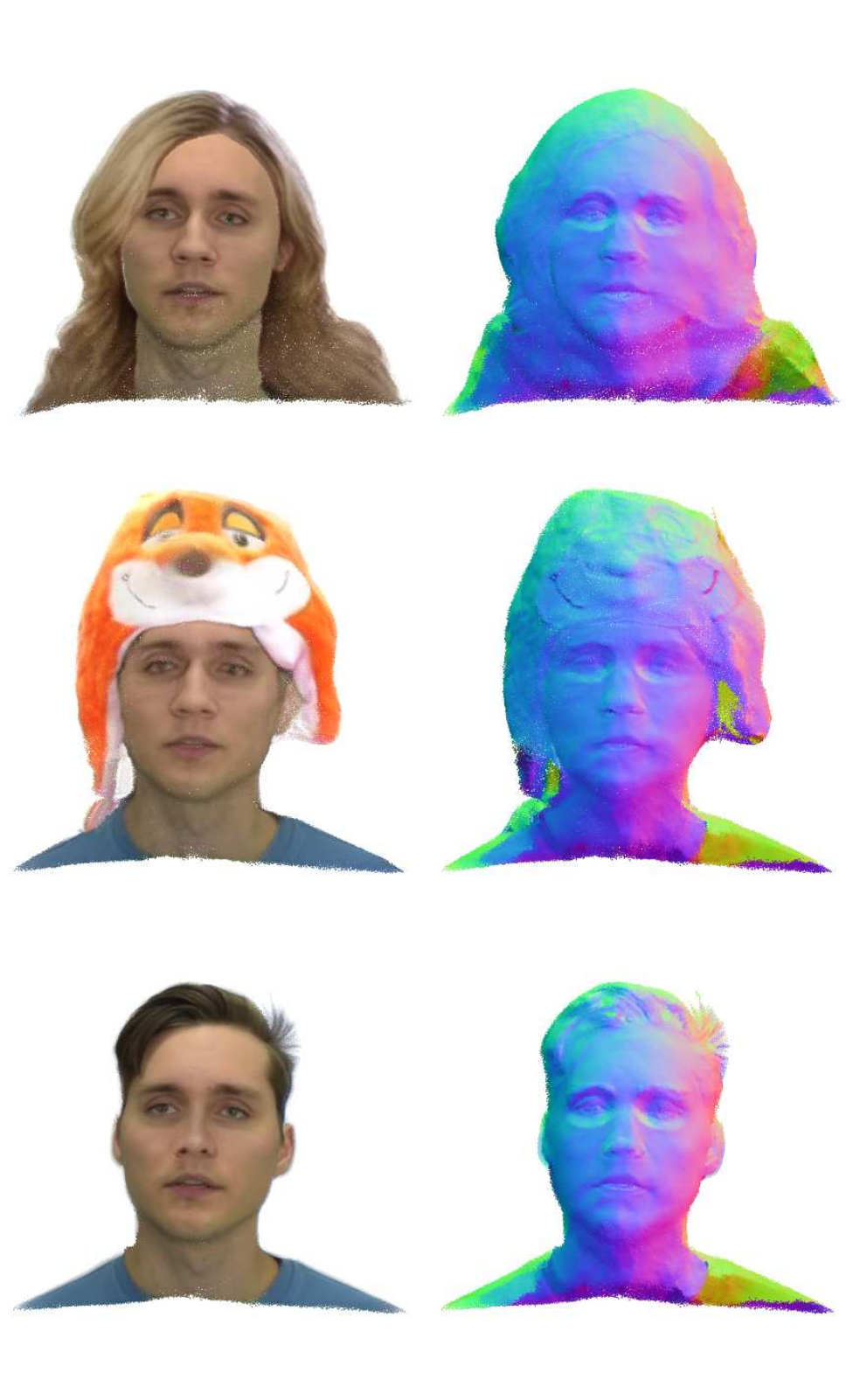}
         \caption{With Normal Loss}
         \label{fig:face_after_resaple}
     \end{subfigure}
     \hfill
     \caption{\textbf{Ablation: normal loss.} The normal loss improves the RGB and geometry quality of our model.}
\label{fig:ablation_normal_loss}
\end{figure}
\noindent \textbf{Normal Loss.}
We show the advantage of our normal loss, which is used for training the avatar model. In \figref{fig:ablation_normal_loss}, the result shows that the normal loss improves the RGB and normal qualities, resulting in more realistic appearances. 

\begin{table}[t]
\vspace{-5px}
\centering
\scriptsize{
\resizebox{\columnwidth}{!}{
\begin{tabular}{l|cc|cc}
\toprule
Method & TL-ID${\uparrow}$ & TG-ID${\uparrow}$ & $L_2{\downarrow}$ & LPIPS${\downarrow}$ \\
\midrule
CD + PA & \textbf{0.9968} & 0.9095 & 114.73 & 0.1428 \\
E4S + PA & 0.9960 & 0.8677 & 110.66 & 0.1234 \\
DELTA & 0.9404 & 0.8368 & 135.03 & 0.1663 \\
Ours$_\text{swap}$+PA & 0.9940 & \textbf{0.9558} & \textbf{97.170} & \textbf{0.1115} \\
\midrule
Ours$_\text{person-gen}$ & \textbf{0.9910} & \textbf{0.9254} & \textbf{105.65} & \textbf{0.1224} \\
Ours$_\text{zero shot}$ & 0.9908 & 0.9178 & 114.38 & 0.1328 \\
\bottomrule
\end{tabular}
}
}
\vspace{-5px}
\caption{\textbf{Temporal metric (TL-ID, TG-ID)~\cite{tzaban2022stitch} and transfer accuracy ($L_2$, LPIPS).} We evaluate the SOTA baselines the same as \tabref{tab:quantiative_hair} with two types of metrics: temporal consistency and preservation of transferred attributes.}
\vspace{-5px}
\label{tab:rebuttal_temporal_transfer}
\end{table}

\begin{figure}[t]
\includegraphics[trim={0 0 0 0},clip,width=1.0\columnwidth]{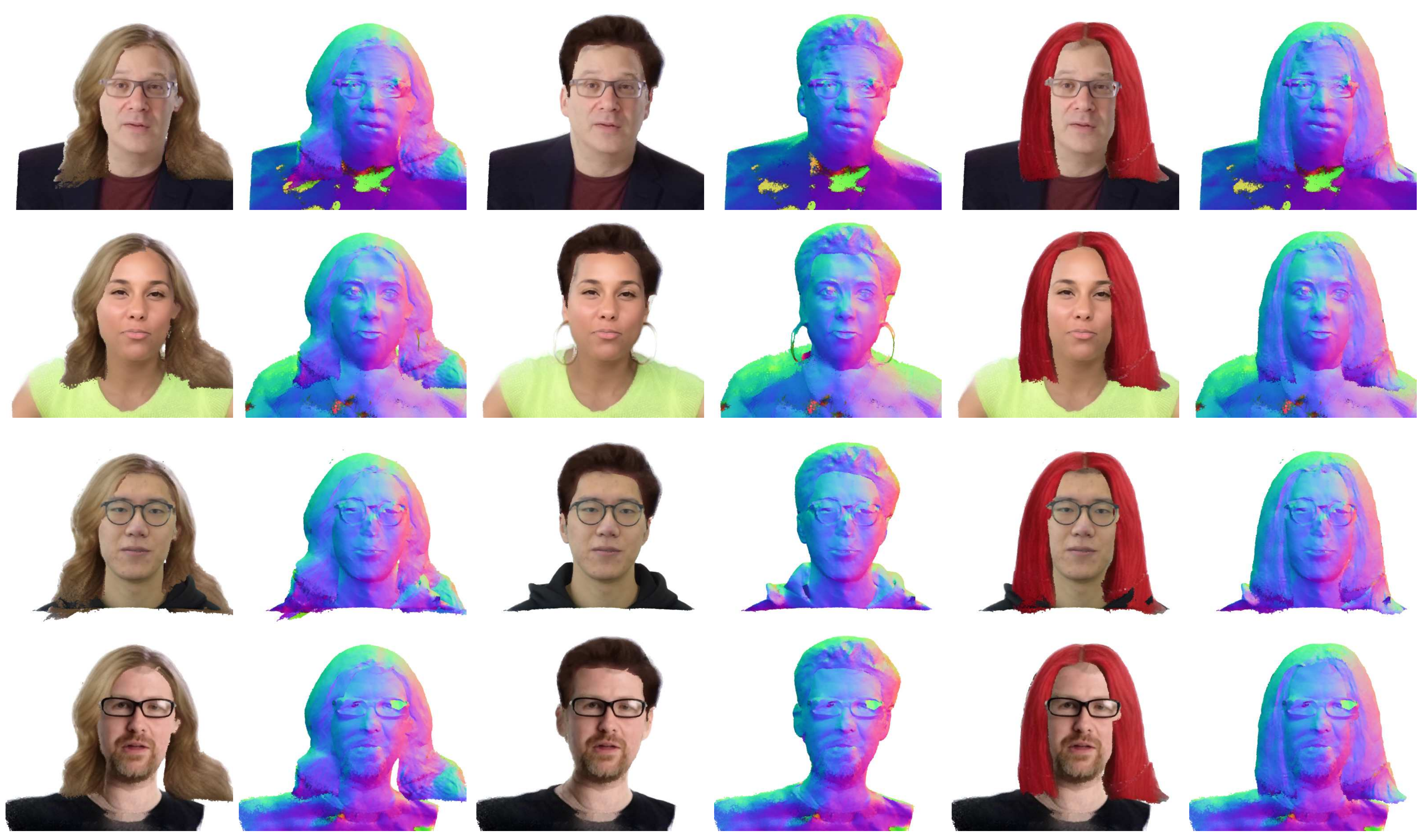}
\caption{\textbf{Additional Results of Zero-Shot Transfer.} PEGASUS robustly transfers facial attributes to any target human without the need for additional training.}
\label{fig:zero_shot_transfer_add}
\end{figure}
\begin{figure}[t]
\includegraphics[trim={0 0 0 0},clip,width=1.0\columnwidth]{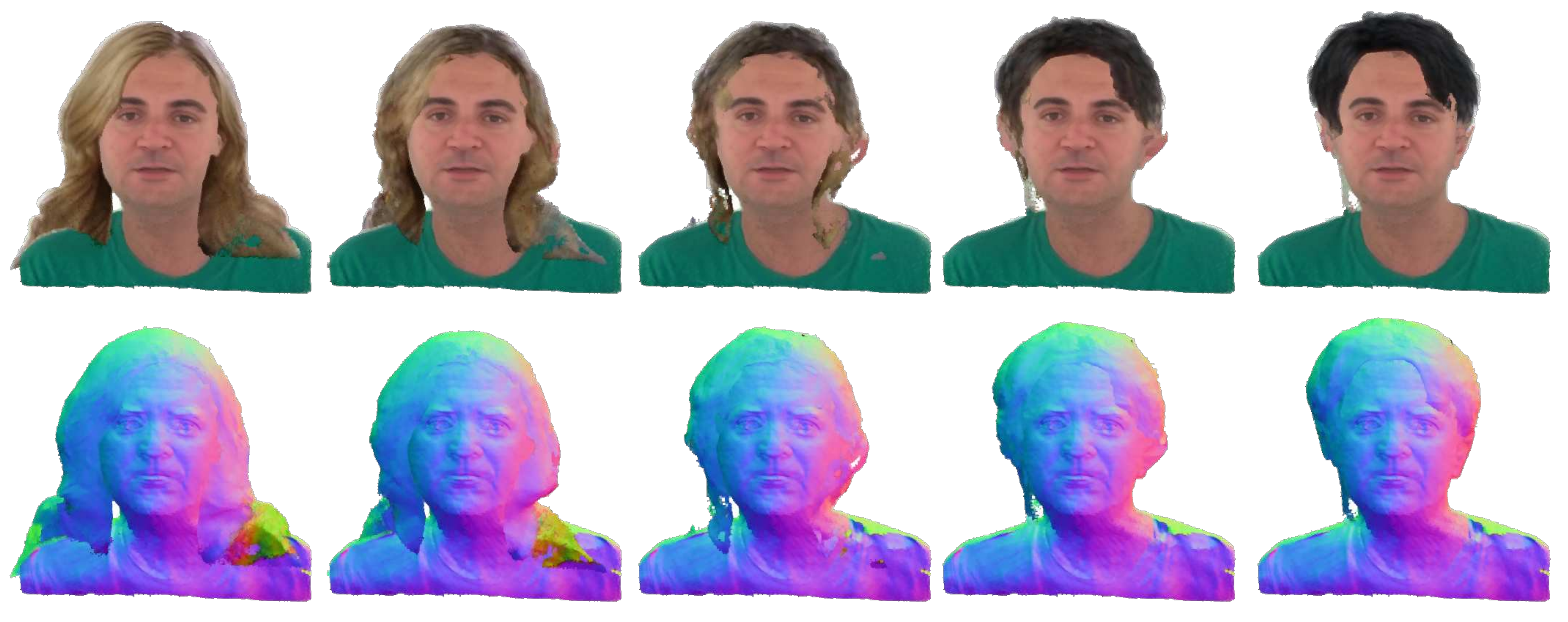}
\caption{\textbf{Zero-Shot Interpolation.} With the help of interpolation-capable segmentation cues by the segmentation network of canonical MLP, we create interpolation in a zero-shot model.}
\label{fig:zero_shot_interpolation}
\end{figure}

\section{Additional Experiments}
\noindent \textbf{Temporal Consistency.}
We employ the temporal consistent metrics~\cite{tzaban2022stitch} to evaluate the preservation of identity across generated image sequences. TL-ID denotes the temporally local identity preservation, which evaluates the consistency of image sequences locally, focusing on the pairs of adjacent frames. TG-ID represents the temporally global identity preservation metric to measure the similarity across all possible pairs of video frames, including those that are not adjacent. We evaluate the same baselines in \tabref{tab:quantiative_hair}. As shown in \tabref{tab:rebuttal_temporal_transfer}, our synthesis method, Ours$_\text{swap}$+PA, represents the best quality of the TG-ID. TL-ID does not show significant differences and performs well across all baselines, as it evaluates consistency only for adjacent frames.

\noindent \textbf{Attributes Transfer.}
In \tabref{tab:rebuttal_temporal_transfer}, we further quantify the preservation of the transferred attributes. We evaluate via the metrics by masking the region except the target attribute. We conduct a comparison between the images generated and the attributes of novel head poses. Our method, Ours$_\text{swap}$+PA, outperforms the SOTA baselines.

\begin{figure}
\centering
    \begin{subfigure}[t]{0.23\columnwidth} 
         \centering
         \includegraphics[width=\columnwidth]{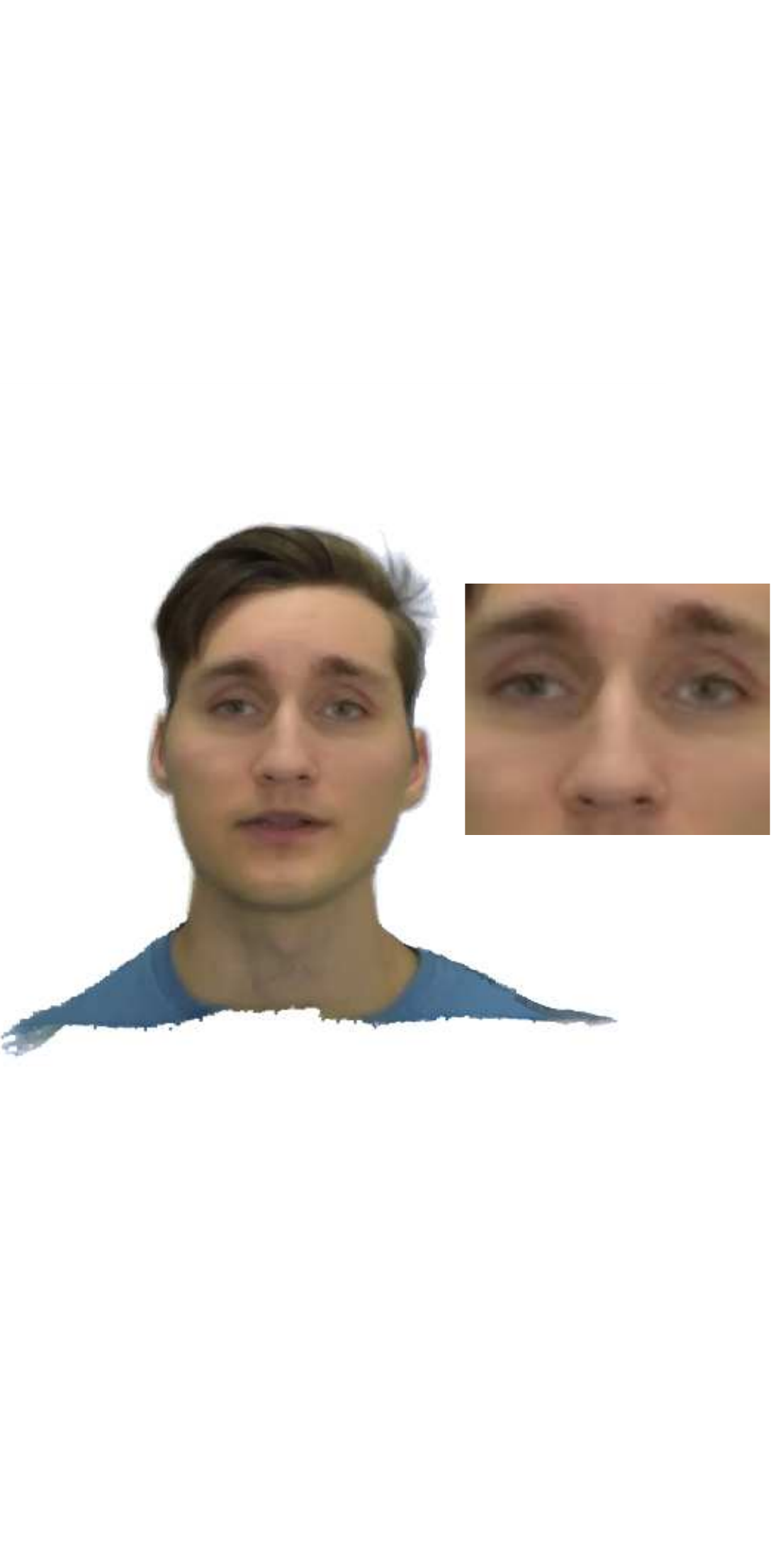}
         \caption{Default}
         \label{fig:random_sample_default}
     \end{subfigure}
     \begin{subfigure}[t]{0.23\columnwidth}
         \centering
         \includegraphics[width=\columnwidth]{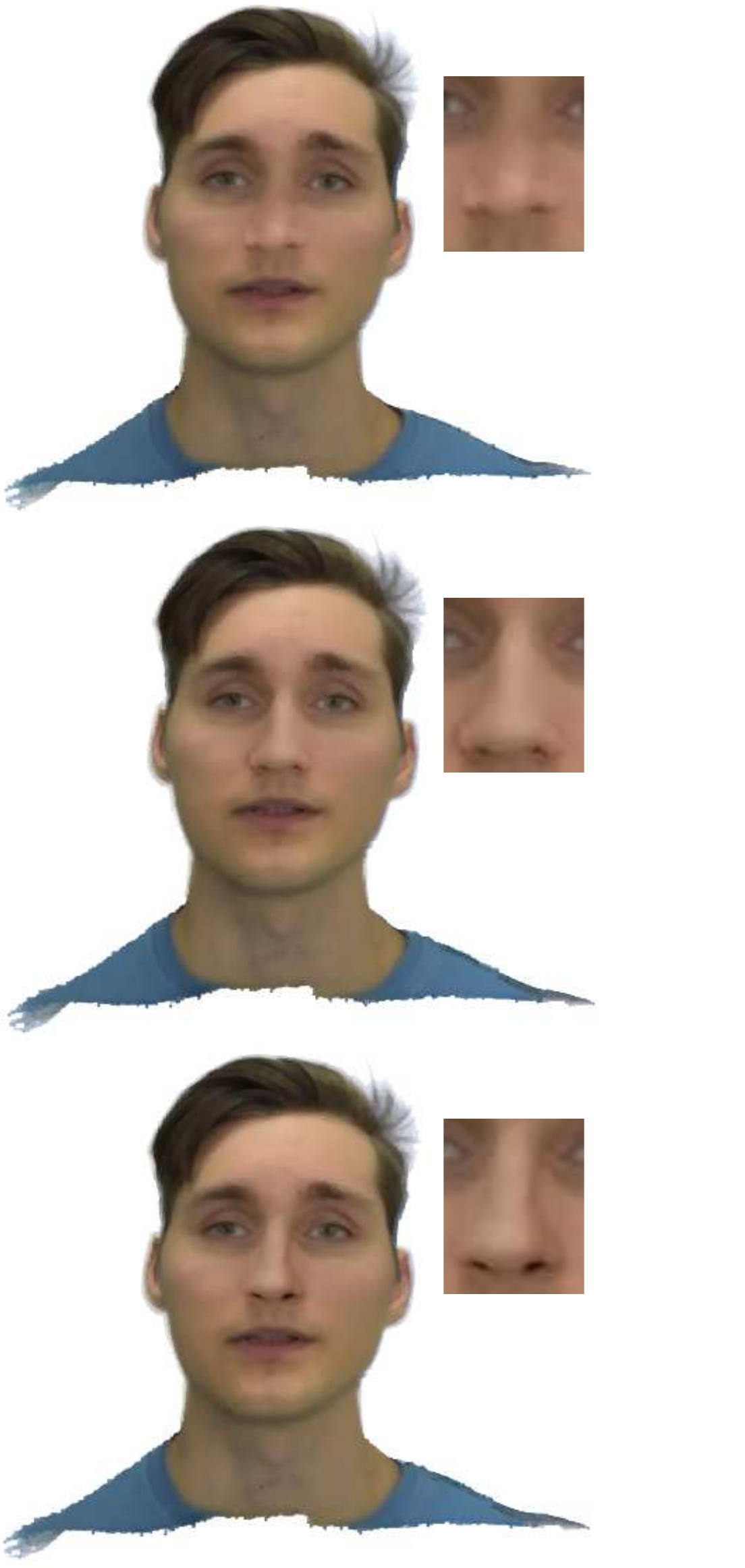}
         \caption{Nose}
         \label{fig:random_sample_nose}
     \end{subfigure}
     \begin{subfigure}[t]{0.22\columnwidth} 
         \centering
         \includegraphics[width=\columnwidth]{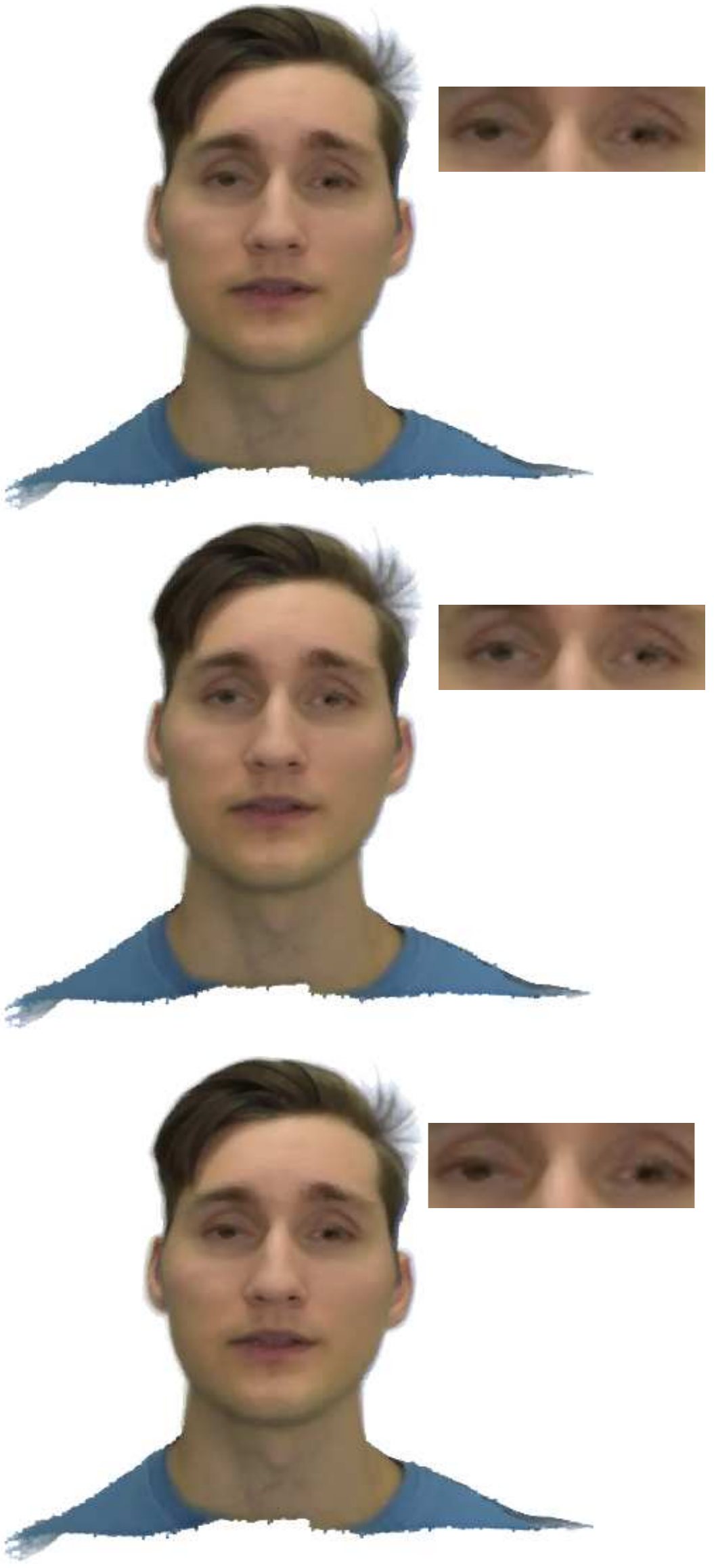}
         \caption{Eyes}
         \label{fig:random_sample_eyes}
     \end{subfigure}
     \begin{subfigure}[t]{0.238\columnwidth}
         \centering
         \includegraphics[width=\columnwidth]{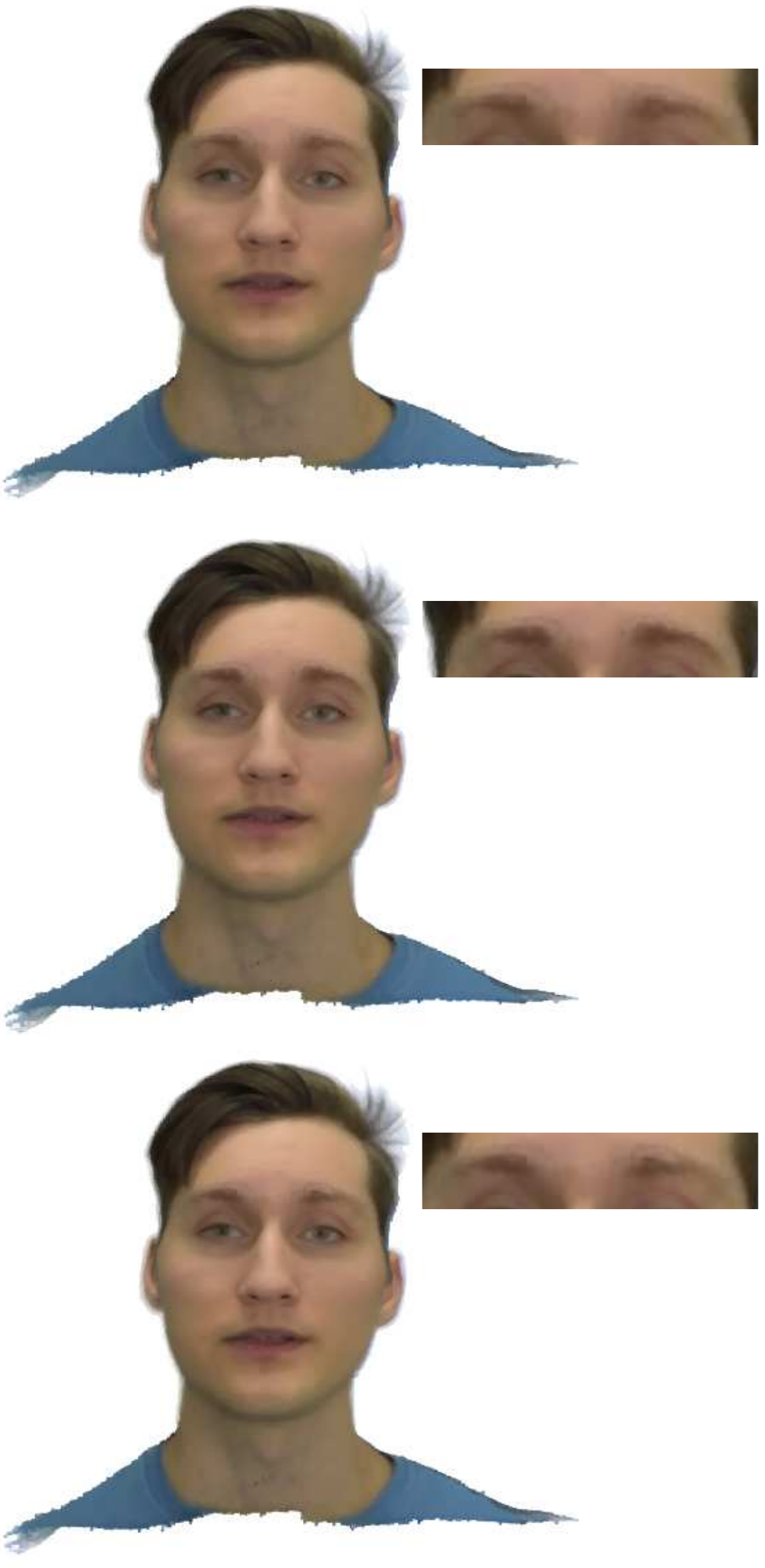}
         \caption{Eyebrows}
         \label{fig:random_sample_eyebrows}
     \end{subfigure}
     \hfill
     \caption{\textbf{Random sampling of PEGASUS.} We randomly sample the latent codes of the PEGASUS.}
\label{fig:random_sampling}
\end{figure}
\begin{figure}[t]
\includegraphics[trim={0 0 0 0},clip,width=1.0\columnwidth]{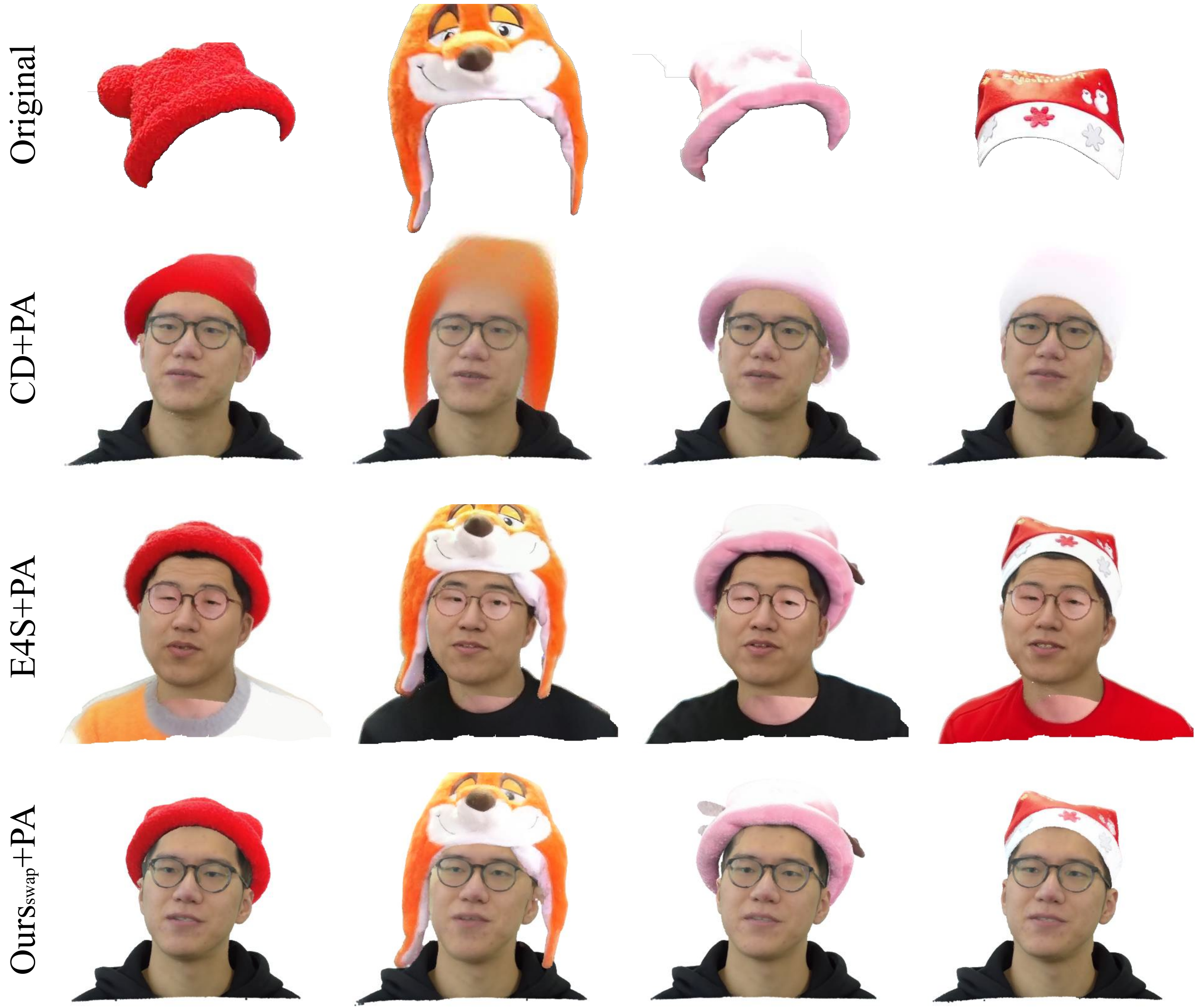}
\caption{\textbf{Single Part-Swapped Avatar on Hat.} Our synthesis method creates high-quality and properly wearing avatars.}
\label{fig:qualitative_results_baselines_hat}
\end{figure}
\begin{figure}[t]
\includegraphics[trim={0 0 0 0},clip,width=\columnwidth]{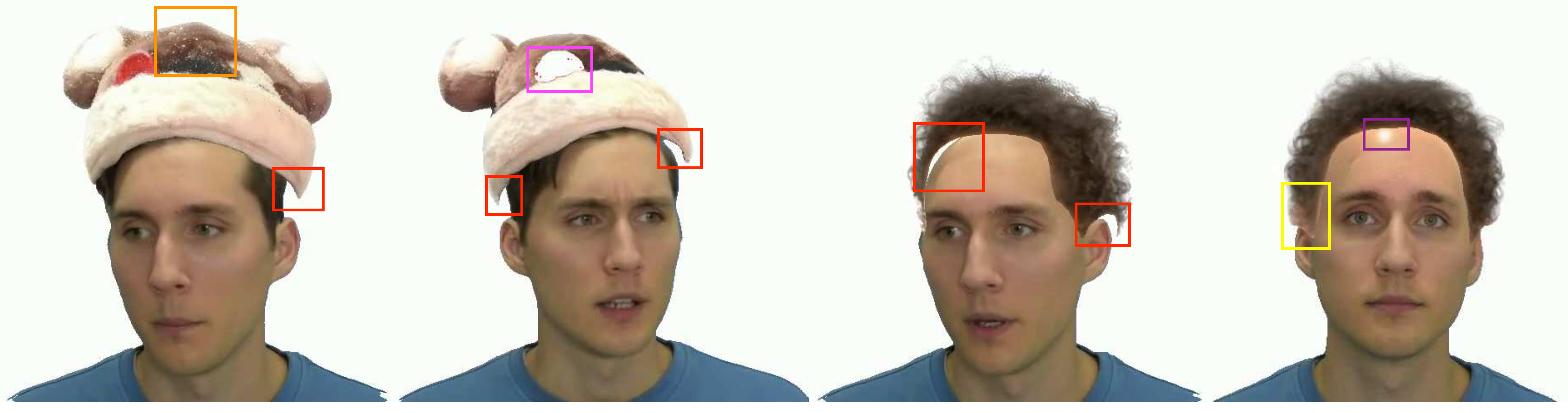}
\caption{\textbf{Limitations of the Synthetic DB.} 
In the synthetic DB generation process, we illustrate the instances of failure cases through color-coded annotations. \textcolor{orange}{Orange box} represents the artifacts occurring during the generation of attribute image. \textcolor{red}{Red box} indicates instances of facial attributes that are physically inconsistent, revealing inaccuracies in the appearance of the face. \textcolor{magenta}{Magenta box} marks the failure case of the segmentation. \textcolor{purple}{Purple box} identifies instances where the diffusion model fails to generate bald faces inconsistently. \textcolor{yellow}{Yellow box} signifies the failure cases in post-processing.}
\label{fig:limitations}
\end{figure}

\section{More Results}
\noindent \textbf{Zero-Shot Transfer.}
\figref{fig:zero_shot_transfer_add} presents additional results of zero-shot transfer. PEGASUS robustly and naturally transfers facial attributes to any target human in the wild. \figref{fig:zero_shot_interpolation} demonstrates facial attribute interpolation in zero-shot modeling, aided by latent code $\mathbf{z}$ interpolation. This shows that segmentation cues are capable of interpolation by the canonical MLP.

\noindent \textbf{Random Sampling.}
We present PEGASUS' latent random sampling result to support the additional generative aspect of our approach in \figref{fig:random_sampling}. We sample each latent code from the Gaussian distribution with the mean and variance of latent codes of each category. As depicted in \figref{fig:random_sampling}, our method successfully generates random samples exhibiting distinguishable facial attributes.

\noindent \textbf{Additional Categories.}
In \figref{fig:qualitative_results_baselines_hat}, our synthesis method maintains the identity better than other baselines and also shows the hat similar to the original while being appropriately worn by the avatar. 

\section{Limitations}
As a limitation, the quality of our personalized avatar still does not reach the photo-realistic quality, showing noticeable artifacts. 
Also, due to the reliance on non-physical-based methods for generating the synthetic DB, our approach exhibits limitations in achieving physical accuracy. We describe the failure cases and limitations of the synthetic DB generation in \figref{fig:limitations}.
 \fi

\end{document}